\documentclass[10pt,twocolumn,letterpaper]{article}

\usepackage[pagenumbers]{cvpr} %

\usepackage{graphicx}
\usepackage{amsmath}
\usepackage{amssymb}
\usepackage{booktabs}
\usepackage{algorithm}
\usepackage{algorithmicx}
\usepackage{algpseudocode}  
\usepackage{color}
\usepackage{tabularray}
\usepackage[table]{xcolor}
\usepackage{multirow}
\usepackage{colortbl}
\usepackage{booktabs}
\usepackage{adjustbox}
\usepackage{diagbox}
\usepackage{booktabs}
\usepackage{xcolor}
\usepackage{enumitem}
\usepackage{balance}

\newcommand{\best}[1]{\textbf{\textcolor{red}{#1}}}

\newcommand{\second}[1]{\underline{\textcolor{blue}{#1}}}
\usepackage{diagbox}
\usepackage{caption}
\usepackage[pagebackref,breaklinks,colorlinks]{hyperref}

\usepackage[capitalize]{cleveref}
\crefname{section}{Sec.}{Secs.}
\Crefname{section}{Section}{Sections}
\Crefname{table}{Table}{Tables}
\crefname{table}{Tab.}{Tabs.}

\usepackage{listings}
\usepackage{float}

\definecolor{codebg}{RGB}{248,248,248}
\definecolor{codekw}{RGB}{0,0,150}
\definecolor{codestr}{RGB}{150,0,0}
\definecolor{codecm}{RGB}{0,100,0}
\definecolor{codenum}{RGB}{120,120,120}

\lstdefinestyle{qwenstyle}{
  basicstyle=\ttfamily\footnotesize,
  breaklines=true,
  columns=fullflexible,
  backgroundcolor=\color{codebg},
  frame=single,
  rulecolor=\color{black},
  showstringspaces=false,
  tabsize=2,
  numbers=left,
  numberstyle=\tiny\color{codenum},
  keywordstyle=\color{codekw}\bfseries,
  stringstyle=\color{codestr},
  commentstyle=\color{codecm},
}

\def\cvprPaperID{21795} %
\def\confName{CVPR}
\def\confYear{2026}

\begin{document}

\title{CreativeVR: Diffusion-Prior-Guided Approach for Structure and Motion Restoration in Generative and Real Videos
}

\author{
Tejas Panambur\textsuperscript{1,†} \quad
Ishan Rajendrakumar Dave\textsuperscript{2,†} \quad
Chongjian Ge\textsuperscript{2} \quad
Ersin Yumer\textsuperscript{2} \quad
Xue Bai\textsuperscript{2} \\
\textsuperscript{1}University of Massachusetts Amherst \quad
\textsuperscript{2}Adobe \\
{\tt\small tpanambur@umass.edu, \{idave, cge, yumer, xubai\}@adobe.com} \\
\small\url{https://daveishan.github.io/creativevr-webpage/} \\
\normalsize\textsuperscript{†}\textit{Equal contribution}
}

\maketitle

\begin{abstract}
Modern text-to-video (T2V) diffusion models can synthesize visually compelling clips, yet they remain brittle at fine\mbox{-}scale structure: even state\mbox{-}of\mbox{-}the\mbox{-}art generators often produce distorted faces and hands, warped backgrounds, and temporally inconsistent motion. Such severe structural artifacts also appear in very low\mbox{-}quality real\mbox{-}world videos. Classical video restoration and super\mbox{-}resolution (VR/VSR) methods, in contrast, are tuned for synthetic degradations such as blur and downsampling and tend to stabilize these artifacts rather than repair them, while diffusion\mbox{-}prior restorers are usually trained on photometric noise and offer little control over the trade\mbox{-}off between perceptual quality and fidelity.
We introduce \emph{CreativeVR}, a diffusion\mbox{-}prior\mbox{-}guided video restoration framework for AI\mbox{-}generated (AIGC) and real videos with severe structural and temporal artifacts. Our deep\mbox{-}adapter\mbox{-}based method exposes a single precision knob that controls how strongly the model follows the input, smoothly trading off between precise restoration on standard degradations and stronger structure\mbox{-} and motion\mbox{-}corrective behavior on challenging content. Our key novelty is a temporally coherent degradation module used during training, which applies carefully designed transformations that produce realistic structural failures.
To evaluate AIGC\mbox{-}artifact restoration, we propose the \textit{AIGC54} benchmark with FIQA, semantic and perceptual metrics, and multi\mbox{-}aspect scoring. CreativeVR achieves state\mbox{-}of\mbox{-}the\mbox{-}art results on videos with severe artifacts and performs competitively on standard video restoration benchmarks, while running at practical throughput (\(\sim\!13\) FPS @ 720p on a single 80\,GB A100).
\end{abstract}

\vspace{-3mm}
\section{Introduction}
\begin{figure*}
    \centering
    \includegraphics[width=0.85\linewidth]{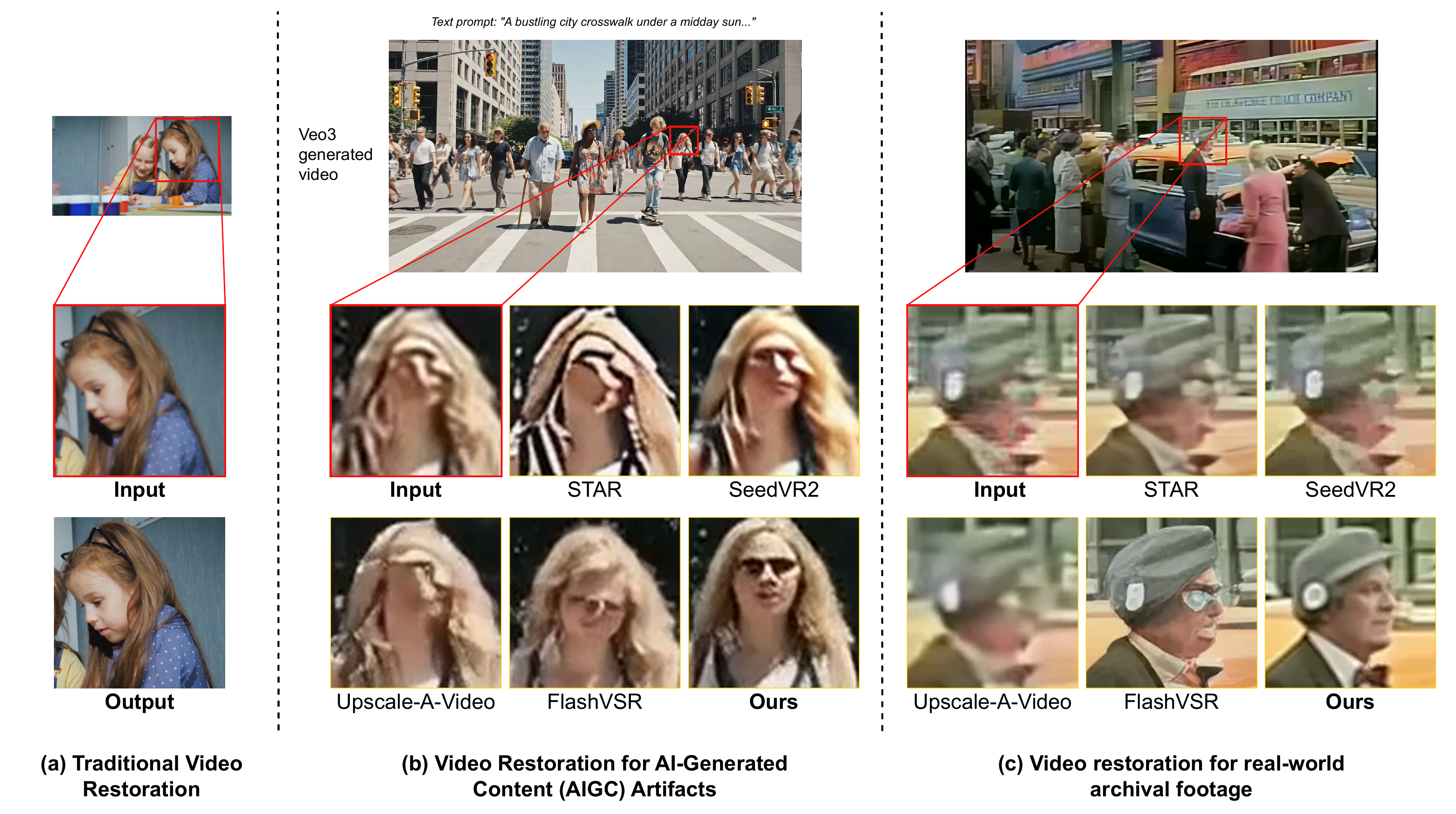}
    \vspace{-3mm}
    \caption{\textbf{Precise Restoration vs Structural-corrective Video Restoration.}
(a) Traditional Video Restoration represents the \emph{precision} regime: recover detail under synthetic degradations when geometry and motion are intact.
(b) AIGC clips exemplify the \emph{prior\mbox{-}guided spatiotemporal correction} regime, the goal is to fix geometric errors and restore temporal coherence while preserving semantics and identity.
(c) Real\mbox{-}world archival footage lies between these extremes and benefits from both precision and prior\mbox{-}guided correction.
Our method modulates this trade\mbox{-}off via a precision knob, behaving precise in case (a) and like a prior\mbox{-}guided corrector in cases (b) and (c).} 
\vspace{-2.5mm}
    \label{fig:teaser}
\end{figure*}

Modern video generation models have made significant advances in generating high\mbox{-}quality video content in terms of semantic composition and adherence to user prompts \cite{google2024veo3,luma2025ray3,openai2025sora2,gao2025seedance}. With the increasing demand for high\mbox{-}quality content, these models are expected to produce production\mbox{-}level videos with high resolution and high frame rates, which requires highly detailed structures and realistic motion dynamics. They are strong overall but brittle at fine\mbox{-}scale structural details. Even cutting\mbox{-}edge models such as Veo3~\cite{google2024veo3}, Ray3~\cite{luma2025ray3}, Kling\cite{kling2025texttovideo}, LTX-2\cite{ltx2_2025} and SORA2~\cite{openai2025sora2} produce structural and temporal artifacts, including imperfect faces and hands, broken topology, warped backgrounds, cross\mbox{-}frame drift, missing object permanence, and fine\mbox{-}detail flicker (Fig.~\ref{fig:teaser}(b)). These artifacts hinder adoption in production, where re\mbox{-}generation or prompt iteration is costly, non\mbox{-}deterministic, and risks deviating from the original intent; this motivates a dedicated \emph{refinement} stage that corrects structure and motion while preserving the semantics and identity of the source video.

Severe artifacts are not unique to AI\mbox{-}generated content (AIGC). Real\mbox{-}world footage, including legacy archives, smartphone videos under compression and low light, low\mbox{-}frame\mbox{-}rate captures, and scans of damaged material, often exhibits degraded detail and temporal instability~\cref{fig:teaser}(c). A practical solution should improve geometry and motion in both \emph{generated} and \emph{real} videos while remaining faithful to the source.

Classical video restoration and super\mbox{-}resolution (VR/VSR) have been highly effective for degradations such as blur, noise, and downsampling, commonly modeled by synthetic or stochastic processes (\cref{fig:teaser}(a)) \cite{wang2019edvr,chan2022basicvsrplusplus}. However, these methods lack semantic priors; when presented with generative artifacts (e.g., a malformed hand or face), they tend to stabilize the artifact rather than repair it. Post\mbox{-}hoc fixes for T2V outputs, such as GAN or diffusion upsamplers and refiners like VideoGigaGAN~\cite{xu2024videogigagan} and VEnhancer~\cite{he2024venhancer}, either alter composition and identity through uncontrolled hallucination or remain domain\mbox{-}specific and fail to generalize across both AI and real footage \cite{xu2024videogigagan,he2024venhancer}. Diffusion\mbox{-}prior restoration methods improve perceptual quality, yet they are typically trained with photometric degradations and lack mechanisms to control the trade\mbox{-}off between \emph{perceptual quality} and \emph{fidelity}. As a result, they may sharpen textures without repairing geometry (Fig.~\ref{fig:teaser}(b)), or over\mbox{-}correct and drift from the source \cite{xie2025star,wang2025seedvr2,xu2024upscaleavideo,zhuang2025flashvsr}.
Diffusion\mbox{-}prior restorers thus sit in an uncomfortable regime: they have strong generative priors but are typically trained on photometric degradations, with no mechanism to steer how much they should trust the corrupted input versus the prior. As a result, they may sharpen textures without repairing geometry (Fig.~\ref{fig:teaser}(b)), or over\mbox{-}correct and drift from the source \cite{xie2025star,wang2025seedvr2,xu2024upscaleavideo,zhuang2025flashvsr}.

In this work we propose \emph{CreativeVR}, a diffusion\mbox{-}prior\mbox{-}guided video restoration framework that explicitly targets this trade\mbox{-}off. We build a deep adapter on top of a frozen text\mbox{-}to\mbox{-}video DiT backbone: the backbone provides a strong generative prior learned from large\mbox{-}scale T2V training, while the adapter is conditioned on the degraded input and trained for restoration. At training time, clean clips and their synthetically degraded counterparts are encoded into a shared latent space, and the adapter features are injected into alternating backbone blocks through a scalar precision control \(\gamma_\ell\). Small \(\gamma_\ell\) values recover the familiar ``precision'' regime of Fig.~\ref{fig:teaser}(a), whereas larger \(\gamma_\ell\) values let the adapter act as a structure\mbox{-} and motion\mbox{-}corrective prior for AIGC and real videos with severe artifacts, as in Fig.~\ref{fig:teaser}(b,c).

A key novelty is our temporally coherent synthetic degradation module, which we use as a \emph{degradation\mbox{-}as\mbox{-}control} curriculum. Instead of training on simple noise or bicubic downsampling, we compose several coherent degradations such as temporal morphing, directional motion blur, grid\mbox{-}based warping, whose parameters evolve smoothly over time. This produces clips with realistic structural failures (e.g., warped faces, wobble, low\mbox{-}FPS blend, and rolling\mbox{-}shutter\mbox{-}like distortions) but without unnatural flicker, and explicitly aligns the diffusion prior toward the hard failure modes observed in modern video generators.

Finally, we evaluate CreativeVR both in the AIGC regime and on standard restoration benchmarks. We curate an \emph{AIGC54} test set from several state\mbox{-}of\mbox{-}the\mbox{-}art video generators and design an evaluation protocol that combines face\mbox{-}structure quality (FIQA), semantic and perceptual metrics, and GPT\mbox{-}based multi\mbox{-}aspect and arena\mbox{-}style preference scoring. Across AIGC clips, CreativeVR attains state\mbox{-}of\mbox{-}the\mbox{-}art results, including up to \textbf{+37\%} relative FIQA improvement over the inputs and consistent gains across all metrics, and it also outperforms existing methods on real videos with severe spatio\mbox{-}temporal artifacts. Our method achieves practical throughput (\(\sim\!13\) FPS @ 720p, \(\sim\!4\) FPS @ 1080p on a single 80\,GB A100) and generalizes in a zero\mbox{-}shot manner up to 1080p resolution.

\noindent Our main contributions are summarized below:
\begin{itemize}[left=0pt, itemsep=0pt, parsep=0pt, topsep=0pt]
    \item We introduce \textbf{CreativeVR}, a diffusion\mbox{-}prior\mbox{-}guided deep adapter on a frozen T2V DiT backbone, equipped with a single precision knob that smoothly trades off between precise restoration and structure\mbox{-}/motion\mbox{-}corrective refinement for both AIGC and real videos.
    \item We propose a \textbf{temporally coherent synthetic degradation} module that composes morphing, directional motion blur, grid\mbox{-}based warping, frame dropping, and spatio\mbox{-}temporal resampling to generate realistic structural artifacts without temporal flicker, providing a targeted training curriculum for diffusion priors.
    \item We establish an \textbf{AIGC\mbox{-}artifact restoration protocol} with the \textit{AIGC54} benchmark and a metric suite combining FIQA and GPT\mbox{-}based evaluation, and demonstrate that CreativeVR achieves state\mbox{-}of\mbox{-}the\mbox{-}art performance on AIGC clips and real videos with severe artifacts, while remaining competitive on standard VR/VSR benchmarks with practical high\mbox{-}resolution throughput.
\end{itemize}

\section{Related Work}

\color{black}
\subsection{Traditional (Precise) Video Restoration}
A large body of image restoration work~\cite{saharia2022image,luo2023image,zhang2021designing,wang2024exploiting,yue2023resshift,li2025diffusion,kong2025deblurdiff,lugmayr2022repaint} has shown that powerful generative priors significantly improve reconstruction under complex degradations. However, video restoration additionally requires \emph{temporal consistency} in both structure and appearance, making frame-independent refiners inadequate for long-range distortions, flicker, and geometric inconsistencies.

Early video restoration (VR) and super\mbox{-}resolution (VSR) extend classical image SR with temporal alignment. EDVR~\cite{wang2019edvr} and TDAN~\cite{tian2020tdan} introduce deformable alignment and multi\mbox{-}frame fusion, while BasicVSR~\cite{chan2022basicvsr} and BasicVSR++~\cite{chan2022basicvsrplusplus} refine this with bidirectional propagation and large temporal receptive fields. RLSP~\cite{fuoli2019efficient} and TMNet~\cite{xu2021temporal} further exploit recurrent memory for long\mbox{-}range consistency. Inspired by image generative models, VideoGigaGAN~\cite{xu2025videogigagan} adapts GigaGAN~\cite{kang2023scaling} to videos by replacing 2D layers with 3D spatiotemporal modules and motion\mbox{-}aware attention. However, these methods are trained under simplified degradations (e.g., bicubic downsampling, Gaussian blur, mild compression) and remain fidelity\mbox{-}oriented, so they do not generalize to complex real\mbox{-}world degradations or the semantic, geometry\mbox{-}level artifacts produced by modern text\mbox{-}to\mbox{-}video models (identity drift, distorted limbs, structural instability). CreativeVR addresses this gap by leveraging a frozen diffusion prior with degradation\mbox{-}conditioned adapters for structure\mbox{-}faithful restoration under both real\mbox{-}world and generative artifacts.

\subsection{Diffusion Models for Video Restoration}
Diffusion-based generative priors provide a more expressive alternative to classical VSR systems by enabling high-quality detail synthesis under complex degradations. Building on image-domain refiners, such as StableSR~\cite{wang2024stablesr}, several works attempt to extend image diffusion upsamplers to videos by introducing temporal consistency modules on top of pretrained image models. For example, STAR~\cite{xie2025star} augments image diffusion backbones with temporal alignment layers to enforce short-range coherence, while VEnhancer~\cite{he2024venhancer} adopts a similar strategy by injecting cross-frame attention to refine geometric structures. With the rapid success of video diffusion models, several recent systems have directly fine-tune high-capacity diffusion models for restoration, of which the representative examples include SeedVR~\cite{wang2025seedvr}, SeedVR2~\cite{wang2025seedvr2}, InfVSR~\cite{zhang2025infvsr}, FlashVideo~\cite{zhang2025flashvideo}, etc. However, these approaches require full fine-tuning of large diffusion transformers, which not only consumes substantial computational resources but also risks altering the pretrained motion and composition priors learned from massive text-video data. In our work, CreativeVR differs by freezing a pretrained text-to-video diffusion transformer and learning lightweight cross-attention adapters conditioned on degraded inputs, achieving efficient and temporally coherent restoration without altering the original backbone.
\vspace{-2mm}
\subsection{Efficient Diffusion Adaptation}
Parameter\mbox{-}efficient fine\mbox{-}tuning (PET) techniques such as Adapters~\cite{houlsby2019adapters} and LoRA~\cite{hu2021lora} enable lightweight specialization of large diffusion models, while ControlNet\mbox{-}style conditioning~\cite{zhang2023adding} has been extended to video. Concat\mbox{-}based models~\cite{ju2025editverse} add separate control encoders and often require full or near\mbox{-}full finetuning, whereas deep adapter approaches such as VACE~\cite{jiang2025vace}, ResTuning~\cite{jiang2023res}, and related low\mbox{-}rank residual controllers provide more modular conditioning with substantially fewer trainable parameters. Building on these ideas, CreativeVR adopts a \emph{degradation\mbox{-}as\mbox{-}control} design: the degraded video itself serves as an internal control signal, injected via lightweight adapters trained under a synthetic degradation curriculum. Unlike external control maps (e.g., edges, depth, poses), this turns adapter\mbox{-}based conditioning into a self\mbox{-}supervised restoration mechanism that corrects both classical degradations and high\mbox{-}level generative artifacts while preserving the motion and composition priors of the frozen video diffusion backbone.

Recent advances in timestep distillation, including CausVid~\cite{yin2025slow}, MMD~\cite{salimans2024multistep}, Self\mbox{-}Forcing~\cite{huang2025self}, and consistency\mbox{-}based distillation~\cite{luo2023latent}, compress 30--50 diffusion steps into only a few sampling steps. CreativeVR is fully compatible with these accelerated backbones: at inference, our adapters can be plugged directly into a distilled base model without additional finetuning, enabling practical high\mbox{-}throughput video restoration.

\begin{figure*}
    \centering
    \includegraphics[width=0.85\linewidth,trim=20 16 25 16,clip]{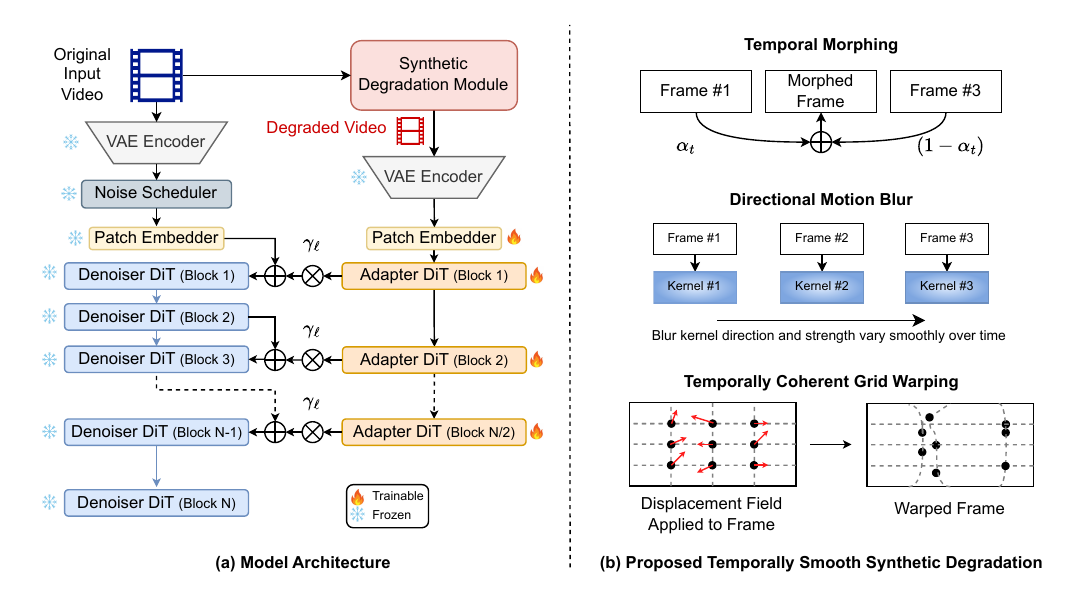}
    \vspace{-2mm}
    \caption{\textbf{CreativeVR overview: diffusion-prior-guided video restoration.}
(a) During training, a clean input clip is passed through a temporally smooth synthetic degradation module to produce a degraded counterpart; both clips are encoded by a frozen VAE and processed by a frozen text-to-video DiT backbone augmented with a lightweight adapter DiT. The adapter is conditioned on degraded latents and injected into alternating backbone blocks via a precision knob \(\gamma_\ell\), allowing the model to trade off prior strength and faithfulness to the input. 
(b) The degradation module composes temporally coherent morphing, directional motion blur, and grid-based warping to mimic realistic structural and motion artifacts, guiding the adapter to learn structure- and motion-corrective restoration behavior.}
\vspace{-2mm}
\label{fig:arch}
\end{figure*}

\section{Method}
\label{sec:method}

We address the problem of diffusion--prior-guided video restoration.  
Given a clean target video \(x \in \mathbb{R}^{T \times H \times W \times 3}\), we synthesize a degraded input video \(\tilde{x}\) and learn a diffusion model that restores \(\tilde{x}\) back to the high--quality space of \(x\). At test time, the model receives only a low--quality video and produces a restored video \(\hat{x}\) at the target resolution.

\subsection{Architecture and Optimization}

Our framework builds on a text-to-video (T2V) diffusion transformer (DiT) backbone, similar to VACE~\cite{jiang2025vace}, augmented with a deep adapter branch (Fig.~\ref{fig:arch}). The T2V DiT backbone is pretrained on a large-scale video generation task and kept frozen during training. The adapter is a lighter DiT with the same block design but only half as many blocks as the backbone.

Let \(E\) denote the video VAE encoder and \(P\) the patch embedder. For each clean training video \(x\), we first construct a degraded counterpart \(\tilde{x} = \mathcal{D}(x; \eta)\), where \(\mathcal{D}\) is our synthetic degradation module with parameters \(\eta\) (Sec.~\ref{subsec:degradation}). Both \(x\) and \(\tilde{x}\) are mapped into the latent space using the same frozen VAE, yielding \(z = E(x)\) and \(\tilde{z} = E(\tilde{x})\). The degraded latent \(\tilde{z}\) is further projected to conditioning tokens \(c = P(\tilde{z})\).
We follow a standard diffusion formulation over the clean latent video \(z\).  
A noise scheduler samples a timestep \(t\) and produces a noisy latent
\setlength{\abovedisplayskip}{4pt}
\setlength{\belowdisplayskip}{4pt}
\begin{equation}
z_t = \alpha_t z + \sigma_t \varepsilon, \qquad \varepsilon \sim \mathcal{N}(0, I),
\end{equation}
which is passed to the frozen DiT denoiser. Let \(h_\ell\) denote the hidden features at the \(\ell\)-th DiT block of the backbone, and let \(a_\ell = A_\ell(c)\) be the output of the corresponding adapter block, which processes the degraded-video tokens \(c\). At a subset of layers \(\ell \in \mathcal{L}_{\text{adapt}}\) we fuse backbone and adapter features via a residual modulation:
\setlength{\abovedisplayskip}{4pt}
\setlength{\belowdisplayskip}{4pt}
\begin{equation}
\label{eq:fusion}
\tilde{h}_\ell = h_\ell + \gamma_\ell\, a_\ell,
\end{equation}
where \(\gamma_\ell \ge 0\) is a scalar \emph{precision knob} that controls the adapter's influence on the backbone. Small values of \(\gamma_\ell\) preserve the frozen T2V prior (strong generative bias, less dependence on degradation), while larger values increase fidelity to the degraded input.
The complete denoiser \(\varepsilon_{\theta,\phi}\) consists of the frozen backbone parameters \(\theta\) and trainable adapter parameters \(\phi\). We optimize only \(\phi\) with the standard noise-prediction diffusion loss
\begin{equation}
\label{eq:diffusion-loss}
\mathcal{L}_{\text{diff}}(\phi)
= \mathbb{E}_{x,t,\varepsilon}\Big[ \big\|
\varepsilon - \varepsilon_{\theta,\phi}(z_t, t, c)
\big\|_2^2 \Big],
\end{equation}
encouraging the coupled backbone--adapter system to denoise towards the clean video latent \(z\) while being guided by the degraded-video tokens \(c\).

\subsection{Synthetic Training Augmentations}
\label{subsec:degradation}
We instantiate the synthetic degradation operator \(\mathcal{D}(x;\eta)\) as a composition of temporally coherent augmentations that mimic natural capture artifacts such as motion blur, geometric wobble, low frame rate, and exposure-induced blending. Given a clean RGB clip \(x = \{X_t\}_{t=1}^T\), the module produces a degraded clip \(\tilde{x} = \mathcal{D}(x;\eta) = \{Y_t\}_{t=1}^T\), with augmentation parameters that evolve smoothly over time to avoid high-frequency flicker.

\noindent\textbf{Structure-preserving spatiotemporal downsampling.}
We jointly downsample the clip in time and space by factors \(s_{\mathrm{temp}}\) and \(s_{\mathrm{spat}}\), then reconstruct it back to the original resolution and frame rate by simple interpolation, mimicking low frame-rate and low-resolution capture while preserving global scene layout.

\noindent\textbf{Temporal morphing.}
We interpolate between adjacent frames to mimic low-shutter or low-FPS blending,
\begin{equation}
    Y_t = \alpha_t X_t + (1 - \alpha_t) X_{t+1},
\end{equation}
where \(\alpha_t \in [\alpha_{\min}, \alpha_{\max}]\) varies smoothly across time.  
This softly blends motion phases across frames, resembling exposure integration in real cameras.

\noindent\textbf{Stochastic frame dropping.}
We sample a binary drop mask \(m_t \sim \mathrm{Bernoulli}(p_{\mathrm{drop}})\) under a maximum run-length constraint and remove frames with \(m_t = 0\); missing frames are then reconstructed by linear interpolation between the nearest retained neighbors.  
This simulates realistic temporal discontinuities observed in mobile or streaming videos while keeping motion trajectories plausible.

\noindent\textbf{Directional motion blur.}
We model exposure-integrated motion by convolving frame with an oriented line kernel,
\begin{equation}
    Y_t = K_t(\theta_t, \ell_t) * X_t,
\end{equation}
where \(\theta_t\) and \(\ell_t\) denote the blur orientation and kernel length, respectively.  
By allowing \(\theta_t\) and \(\ell_t\) to change smoothly over time, we imitate camera or object motion that produces temporally varying, yet coherent, motion blur.

\noindent\textbf{Grid-based spatial warping.}
To simulate rolling-shutter wobble and low-frequency geometric distortions, we generate a displacement field
\(d_t = (d_t^x, d_t^y)\) by upsampling smooth low-resolution noise and apply it via backward warping,
\begin{equation}
    Y_t(u) = X_t\big(u + d_t(u)\big),
\end{equation}
where \(u\) indexes pixel coordinates.  
The resulting elastic, wavy deformations resemble lens wobble, heat haze, or hand-held jitter while keeping local geometry plausible.

\noindent\textbf{Smooth parameter trajectories.}
All temporal parameters \((\alpha_t, \theta_t, \ell_t, d_t)\) are drawn from low-frequency trajectories obtained using sinusoidal or Perlin-noise bases and optionally smoothed with a 1D box filter.  
This design avoids abrupt parameter jumps and prevents synthetic flicker, yielding degradations that evolve smoothly like real-world capture effects.
\subsection{Sampling}
At inference time we discard the clean branch and condition only on a degraded clip \(\tilde{x}\): we encode \(\tilde{x}\) to latents \(\tilde{z} = E(\tilde{x})\), obtain conditioning tokens \(c = P(\tilde{z})\), and run the diffusion sampler on \(\tilde{z}\) to produce a restored latent \(\hat{z}\), which is decoded by the VAE decoder into the final video \(\hat{x}\). For efficient sampling, we replace the frozen T2V backbone \(\theta\) with a timestep-distilled student DiT backbone \(\theta_s\) based on CausVid~\cite{yin2025slow}, while reusing the same trained adapter blocks \(A_\ell\). This plug-and-play swap reduces the number of sampling steps from 50 to 4, markedly improving throughput without degrading restoration quality.

\section{Experiments}

\subsection{Dataset and Implementation Details}

\noindent\textbf{Datasets.} 
We train CreativeVR on the Mixkit split of Open-Sora-Plan v1.1.0~\cite{opensoraplan2024}, which provides $\sim$10K open-source video clips spanning diverse scenes. All  videos are sampled as 49-frame sequences and resized to $480\times832$ before augmentation.
To evaluate video restoration under AIGC artifacts, we propose the \textit{AIGC54} video dataset. Our dataset contains 54 videos collected from Veo3~\cite{google2024veo3}, Pika2.2~\cite{pika2025pika22}, Firefly~\cite{adobe2025firefly}, Ray3~\cite{luma2025ray3}, and Wan2.1~\cite{wan2025wan}. Each video is a 5-second clip covering a wide variety of scenarios, including traditional dance, tourist groups, courtroom debates, crowded streets, sports events, and newsroom environments, capturing challenging motion and structural artifacts.

\begin{table*}
\centering
\small
\begingroup
\arrayrulecolor{gray!50}
\begin{tabular}{l|c|c|c|c|c|c|c|c|c}
\hline
\arrayrulecolor{black}
\toprule
\diagbox[
  width=2.4cm,
  height=1.2cm,
  trim=l,
  font=\scriptsize
]{\textbf{Quality}\\\textbf{Metric}}{\textbf{Upsampling}\\\textbf{Method}} 
& \begin{tabular}[c]{@{}c@{}}\textbf{FlashVSR}\\\cite{zhuang2025flashvsr}\end{tabular} 
& \begin{tabular}[c]{@{}c@{}}\textbf{Real-}\\\textbf{ESRGAN}\\\cite{wang2021real}\end{tabular}
& \begin{tabular}[c]{@{}c@{}}\textbf{Real-}\\\textbf{Viformer}\\\cite{zhang2024realviformer}\end{tabular}
& \begin{tabular}[c]{@{}c@{}}\textbf{ResShift}\\\cite{yue2023resshift}\end{tabular}
& \begin{tabular}[c]{@{}c@{}}\textbf{SeedVR2}\\\cite{wang2025seedvr2}\end{tabular}
& \begin{tabular}[c]{@{}c@{}}\textbf{STAR}\\\cite{xie2025star}\end{tabular}
& \begin{tabular}[c]{@{}c@{}}\textbf{Upscale-A}\\\textbf{-Video}\\\cite{xu2024upscaleavideo}\end{tabular}
& \begin{tabular}[c]{@{}c@{}}\textbf{VEnhancer}\\\cite{he2024venhancer}\end{tabular}
& \begin{tabular}[c]{@{}c@{}}\textbf{Ours}\\\textbf{}\end{tabular}  \\
\midrule
{eDifFIQA}  & 19.60 & -0.60 & 9.40 & 4.50 & 21.60 & \second{28.90} & -6.80 & \second{28.90} & \best{35.60} \\
{DifFIQA}   & -0.40 & 0.40 & 0.10 & -0.10 & 0.40 & \best{1.10} & -0.30 & 0.50 & \second{0.70} \\
{CLIB-FIQA} & 9.20 & -1.70 & 2.70 & 1.10 & 8.90 & 7.00 & -0.50 & \second{10.90} & \best{14.80} \\
{CR-FIQA}   & 3.80 & 6.30 & 0.70 & 3.90 & 6.90 & 12.70 & 0.30 & \second{12.90} & \best{17.20} \\
{MR-FIQA}   & 16.10 & -6.20 & -1.30 & -1.70 & 20.40 & \second{22.50} & -8.10 & 18.60 & \best{37.40} \\
{FaceQAN}   & 2.50 & -2.00 & -8.70 & -7.00 & 13.40 & \best{34.40} & -11.30 & 10.60 & \second{30.60} \\
{Aesthetic Score} & 8.32 & -2.29 & 2.15 & 0.72 & 4.16 & 4.16 & -1.72 & \second{10.22} & \best{11.19} \\
{Objectness} & \second{3.45} & 0.00 & 0.04 & 0.00 & 1.61 & 1.16 & 0.00 & 2.06 & \best{4.24} \\
\bottomrule
\end{tabular}
\arrayrulecolor{black}
\endgroup
\vspace{-3mm}
\caption{\textbf{Structural Integrity Evaluation for AIGC-artifacts} All numbers denote relative improvement (\%) over input videos.}
\end{table*}

\begin{table*}
\centering
\small
\begingroup
\setlength{\tabcolsep}{3pt}
\arrayrulecolor{gray!50}
\begin{tabular}{l|c|c|c|c|c|c|c|c|c}
\hline
\arrayrulecolor{black}
\toprule
\diagbox[
  width=2.4cm,
  height=1.2cm,
  trim=l,
  font=\scriptsize
]{\textbf{Quality}\\\textbf{Metric}}{\textbf{Upsampling}\\\textbf{Method}}
& \begin{tabular}[c]{@{}c@{}}\textbf{FlashVSR}\\\cite{zhuang2025flashvsr}\end{tabular}
& \begin{tabular}[c]{@{}c@{}}\textbf{Real-}\\\textbf{ESRGAN}\\\cite{wang2021real}\end{tabular}
& \begin{tabular}[c]{@{}c@{}}\textbf{Real-}\\\textbf{Viformer}\\\cite{zhang2024realviformer}\end{tabular}
& \begin{tabular}[c]{@{}c@{}}\textbf{ResShift}\\\cite{yue2023resshift}\end{tabular}
& \begin{tabular}[c]{@{}c@{}}\textbf{SeedVR2}\\\cite{wang2025seedvr2}\end{tabular}
& \begin{tabular}[c]{@{}c@{}}\textbf{STAR}\\\cite{xie2025star}\end{tabular}
& \begin{tabular}[c]{@{}c@{}}\textbf{Upscale-A}\\\textbf{-Video}\\\cite{xu2024upscaleavideo}\end{tabular}
& \begin{tabular}[c]{@{}c@{}}\textbf{VEnhancer}\\\cite{he2024venhancer}\end{tabular}
& \textbf{Ours} \\
\midrule
Visual Quality         & \second{7.86}&	6.91&	6.73&	7.05&	7.36&	8.14&	6.73&	7.41&	\best{8.05} \\
Temporal Consistency   &\second{8.05}&	6.82&	7.00&	7.32&	7.32&	8.27&	7.95	&7.91	& \best{8.55} \\
Face Quality           & \best{6.64}&	5.14&	4.95&	5.05&	5.5&	5.77&	4.41	&5.27&	\second{6.23} \\
Motion Realism         & 7.95&	6.68&	7.18&	7.09&	7.14&	\second{8.05}&	7.68&	7.41&	\best{8.27} \\
Lighting \& Atmosphere & \best{8.55}&	8.05&	8.05&	8.36&	8.23&	\second{8.5}&	7.82&	8.36&	8.04 \\
Detail Preservation    & \second{7.73}&	6.55&	6.23&	6.73&	7.00&	\second{7.73}&	6.45&	6.95&	\best{7.82} \\
\hline
Overall Mean           & \second{7.80}&	6.69&	6.69&	6.93&	7.10&	7.74&	6.84&	7.22&	\best{7.82} \\
\bottomrule
\end{tabular}
\arrayrulecolor{black}
\endgroup
\vspace{-3mm}
\caption{\textbf{Multi-Aspect Video Quality Assessment for AIGC-artifacts} Higher is better.}
\vspace{-3mm}

\label{tab:creativity_benchmark}
\end{table*}

\begin{table*}[t]
\centering
\small
\begingroup
\setlength{\tabcolsep}{3pt}
\arrayrulecolor{gray!50}
\resizebox{\linewidth}{!}{%
\begin{tabular}{l l|
c |c| c| c| c| c| c| c| c| c}
\arrayrulecolor{black}
\toprule
\textbf{Dataset} & \textbf{Metrics}
& \begin{tabular}[c]{@{}c@{}}\textbf{FlashVSR}\\\cite{zhuang2025flashvsr}\end{tabular}
& \begin{tabular}[c]{@{}c@{}}\textbf{Real-}\\\textbf{ESRGAN}\\\cite{wang2021real}\end{tabular}
& \begin{tabular}[c]{@{}c@{}}\textbf{Real-}\\\textbf{Viformer}\\\cite{zhang2024realviformer}\end{tabular}
& \begin{tabular}[c]{@{}c@{}}\textbf{ResShift}\\\cite{yue2023resshift}\end{tabular}
& \begin{tabular}[c]{@{}c@{}}\textbf{SeedVR2}\\\cite{wang2025seedvr2}\end{tabular}
& \begin{tabular}[c]{@{}c@{}}\textbf{STAR}\\\cite{xie2025star}\end{tabular}
& \begin{tabular}[c]{@{}c@{}}\textbf{Upscale-A}\\\textbf{-Video}\\\cite{xu2024upscaleavideo}\end{tabular}
& \begin{tabular}[c]{@{}c@{}}\textbf{VEnhancer}\\\cite{he2024venhancer}\end{tabular}
& \begin{tabular}[c]{@{}c@{}}\textbf{Ours}\\\textbf{(Medium)}\end{tabular}
& \begin{tabular}[c]{@{}c@{}}\textbf{Ours}\\\textbf{(Strong)}\end{tabular} \\

\midrule

\multirow{4}{*}{\textbf{UDM10}\cite{tao2017detail}}
& PSNR $\uparrow$ & 25.212 & 29.359 & \second{29.561} & 28.944 & 28.634 & 28.335 & 28.124 & 25.308 & \best{29.680} & 27.350 \\
& SSIM $\uparrow$ & 0.532  & 0.855  & \second{0.864}  & 0.839  & 0.843  & 0.838  & 0.821  & 0.784  & \best{0.871}  & 0.836  \\
& LPIPS $\downarrow$ & 0.473 & 0.251 & \second{0.189} & 0.211 & 0.229 & \second{0.189} & 0.242 & 0.288 & \best{0.135} & 0.234 \\
& DISTS $\downarrow$ & 0.141 & 0.151 & 0.111 & 0.107 & 0.112 & \second{0.095} & 0.129 & 0.141 & \best{0.069} & 0.111 \\
\midrule

\multirow{4}{*}{\textbf{SPMCS}\cite{yi2019progressive}}
& PSNR $\uparrow$ & \best{26.405} & 20.089 & 20.092 & 19.382 & 19.147 & 18.130 & 19.440 & 19.272 & \second{26.286} & 26.112 \\
& SSIM $\uparrow$ & 0.389  & 0.540  & 0.493  & 0.492  & 0.484  & 0.446  & 0.494  & 0.507  & \best{0.854}  & \second{0.849}  \\
& LPIPS $\downarrow$ & 0.353 & 0.365 & 0.229 & 0.234 & 0.196 & 0.304 & 0.289 & 0.345 & \best{0.133} & \second{0.144} \\
& DISTS $\downarrow$ & 0.151 & 0.209 & 0.135 & 0.132 & 0.104 & 0.153 & 0.157 & 0.167 & \best{0.066} & \second{0.069} \\
\midrule

\multirow{4}{*}{\textbf{REDS30}\cite{nah2019ntire}}
& PSNR $\uparrow$ & 25.566 & 25.527 & \second{26.146} & 24.474 & \best{26.380} & 20.918 & 24.898 & 22.879 & 26.022 & 26.036 \\
& SSIM $\uparrow$ & 0.397  & 0.727  & 0.760  & 0.677  & \best{0.783}  & 0.594  & 0.684  & 0.643  & \second{0.779}  & 0.776  \\
& LPIPS $\downarrow$ & 0.389 & 0.359 & \second{0.159} & 0.231 & \second{0.159} & 0.284 & 0.240 & 0.356 & \second{0.159} & \best{0.139} \\
& DISTS $\downarrow$ & 0.115 & 0.161 & 0.080 & 0.105 & 0.077 & 0.132 & 0.113 & 0.138 & \second{0.074} & \best{0.060} \\
\bottomrule
\end{tabular}}
\arrayrulecolor{black}
\endgroup
\vspace{-4mm}
\caption{\textbf{Reference-based benchmarks across datasets.} Method columns appear in alphabetical order. $\uparrow$ higher is better, $\downarrow$ lower is better.}
\vspace{-2mm}
\label{tab:ref_benchmarks}
\end{table*}

\noindent\textbf{Evaluation Metrics}
\noindent\textbf{Evaluation metrics.} To assess AIGC artifacts on faces, we score face crops with six Face Image Quality Assessment (FIQA) models: eDifFIQA~\cite{babnik2024ediffiqa}, DifFIQA~\cite{babnik2023diffiqa}, CLIB\mbox{-}FIQA~\cite{ou2024clib},  CR\mbox{-}FIQA~\cite{boutros2023cr}, MR\mbox{-}FIQA~\cite{ou2025mr}, and FaceQAN~\cite{babnik2022faceqan}. We additionally report an aesthetic score from a CLIP-LAION initialized model and the objectness confidence of a YOLOv8\mbox{-}based face detector to track perceptual appeal and detection reliability.

\noindent\textbf{Implementation Details.}
We utilize the WAN2.1-based 1.3B-parameter DiT model as the base architecture, with deep adapters inserted into alternating DiT blocks. We train on $T = 49$ frames resized to $480 \times 832$ (divisible by latent strides). Training is fast and converges in approximately 5k iterations on 8 NVIDIA H100 GPUs, with a batch size of 1 per GPU. We use the AdamW optimizer with a learning rate of $10^{-4}$, gradient accumulation $\times 2$, and global gradient norm clipping at $2.0$. The parameter $\gamma_\ell$ is initialized to $1.0$ during training.

During sampling, we replace the teacher backbone weights with the student CausVid model and use 4 sampling steps. By default, the parameter $\gamma_\ell$ is set to $0.4$ for prior-guided restoration and $1.0$ for precise video restoration benchmarks. Our method supports zero-shot inference for input resolutions ranging from $176$p up to $1080$p, achieving approximately $13$ FPS at $720$p and $4$ FPS at $1080$p on a single 80\,GB A100 GPU.

\begin{figure*}
    \centering
    \includegraphics[width=\linewidth]{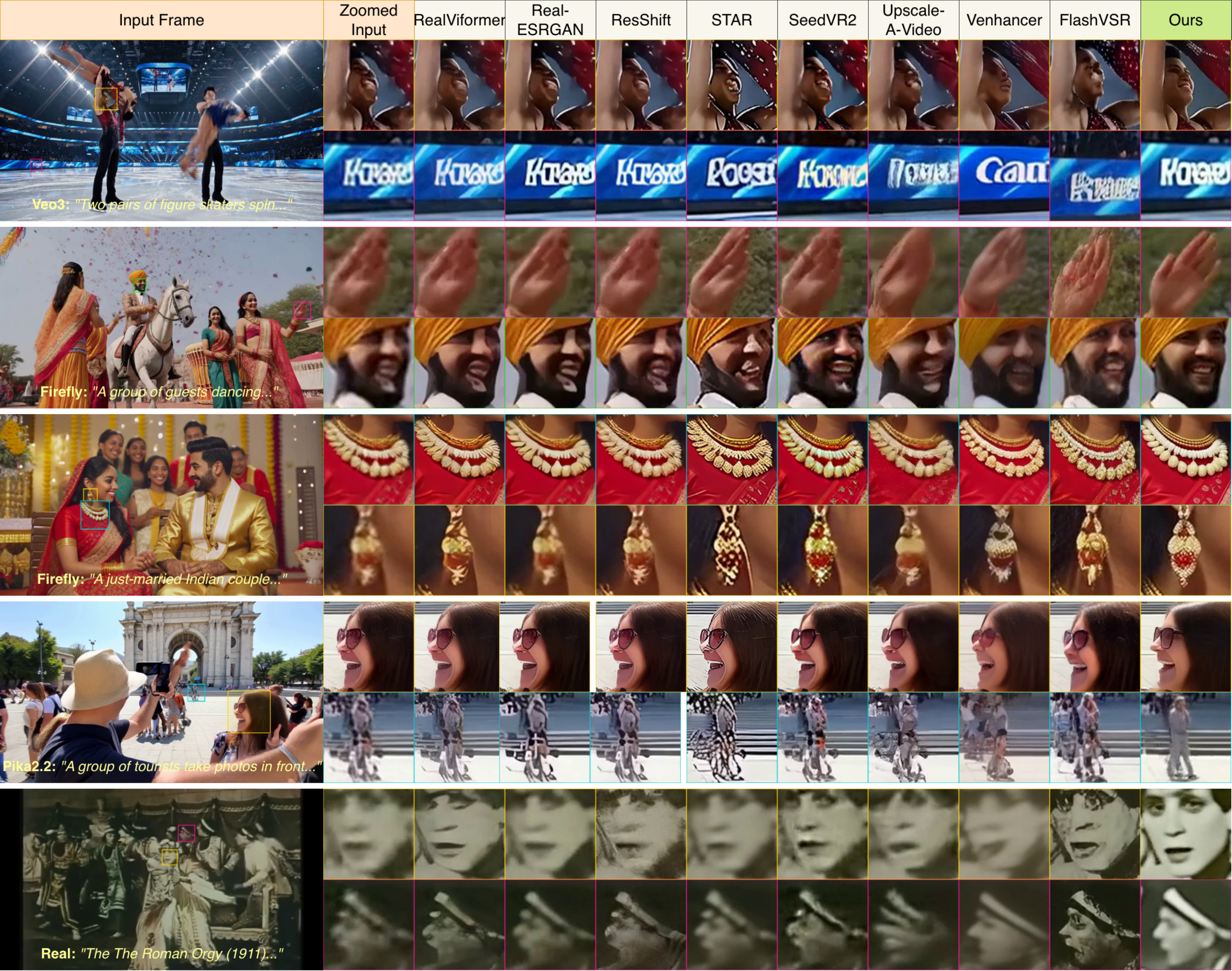}
    \vspace{-7mm}
    \caption{Qualitative Results. We comprehensively compare our method against a wide range of video restoration competitors.} 
    \vspace{-5mm}
    \label{fig:AIGC_Qual}
\end{figure*}
\subsection{Quantitative Comparison}

We consider two main set of evaluation based on prior guided corrective which requires severe structural-motion artifact correction and precision-based traditional video restoration. 
\vspace{-3mm}
\subsubsection{Prior-guided corrective Video Restoration.}
\noindent\textbf{Structural Integrity Evaluation.}
We quantitatively assess AIGC artifacts on the \textit{AIGC54} dataset and compare against state-of-the-art video restoration methods:
FlashVSR~\cite{zhuang2025flashvsr}, Real-ESRGAN~\cite{wang2021real}, Real-Viformer~\cite{zhang2024realviformer}, ResShift~\cite{yue2023resshift}, SeedVR2~\cite{wang2025seedvr2}, STAR~\cite{xie2025star}, Upscale-A-Video~\cite{xu2024upscaleavideo}, and VEnhancer~\cite{he2024venhancer}.
Because generated videos lack paired references, we evaluate with no-reference, face-centric quality signals that are sensitive to structural plausibility.
Concretely, we uniformly sample 16 frames per video, detect faces on the \emph{input} using a YOLOv8-based face detector, expand boxes by 10\% on each side, and extract the identical boxes from each restored output (no re-detection) to avoid selection bias.
We report per-crop \emph{relative gain} over the input, i.e., $\Delta\!=\!\mathrm{score}_{\text{out}}-\mathrm{score}_{\text{in}}$, averaged over crops and frames (see Table~\ref{tab:creativity_benchmark}).

To target structural integrity, we adopt six FIQA metrics: diffusion-prior robustness (DifFIQA/eDifFIQA~\cite{babnik2023diffiqa,babnik2024ediffiqa}), recognition-embedding separability and confidence calibration (CR-FIQA/CLIB-FIQA~\cite{boutros2023cr,ou2024clib}), and multi-reference plus adversarial-noise sensitivity (MR-FIQA/FaceQAN~\cite{ou2025mr,babnik2022faceqan}); together, these provide reliable signals of geometric/edge correctness, texture realism, and identity preservation under AIGC artifacts.
Notably, sharpening-oriented baselines sometimes fail to improve face quality (or even regress) because they enhance high-frequency detail without correcting warped geometry; in contrast, diffusion-prior-guided approaches tend to yield consistent positive $\Delta$ on FIQA.
Our method achieves the best gains over FIQA up to \textbf{+37\%}, indicating strong structural correction.

Beyond FIQA, we also report an aesthetic score from a CLIP-LAION-initialized model which captures overall perceptual appeal of the crops and the objectness confidence of the YOLOv8 face detector as a proxy for semantic reliability.
Our method attains the highest aesthetic gain and the largest increase in detection confidence, corroborating that structural corrections translate to perceptually cleaner and more reliably recognized faces.

\noindent\textbf{Multi-Aspect Video Quality Assessment} To complement structural metrics, we employ a GPT-based evaluator that scores each clip along six perceptual dimensions: visual quality, temporal consistency, face quality, motion realism, lighting and atmosphere, and detail preservation. Scores follow a 0--10 scale and are averaged to obtain an overall quality measure. This protocol captures perceptual factors that extend beyond distortion-oriented metrics. Table~\ref{tab:creativity_benchmark} summarizes the results across all baselines. Our method attains the highest mean score and shows consistent improvements across temporal coherence, visual quality, and detail preservation, indicating the effectiveness of diffusion priors in correcting structure while maintaining scene semantics.
To verify reliability, four independent users evaluated 10 random test videos (\(\approx 20\%\) of the AIGC dataset), and GPT's rankings showed strong qualitative agreement with majority human preference. Besides, we  employ a pairwise arena-style evaluation using a GPT judge for a further comparison, of which the results are shown in the appendix.

\vspace{-2mm}
\subsubsection{Traditional Video Restoration.}
On standard video restoration benchmarks, CreativeVR attains PSNR/SSIM comparable to strong SR/VSR baselines and competitive LPIPS/DISTS scores (Table.~\ref{tab:ref_benchmarks}). 
This shows that a model optimized for challenging AIGC and real-world artifacts also transfers well to classical degradations without task-specific tuning.

\subsection{Qualitative Comparison}
Fig.~\ref{fig:AIGC_Qual} shows cases where prior methods fail to remove AIGC or real-world artifacts with blurred or fused facial profiles, smeared hands and jewelry, and distorted signage. CreativeVR restores clean, geometrically plausible faces, fingers, and fine details while preserving pose and scene layout. More examples are in the supplementary material.

\subsection{Ablation Study}
\begin{figure}[t]
    \centering
    \includegraphics[width=\linewidth]{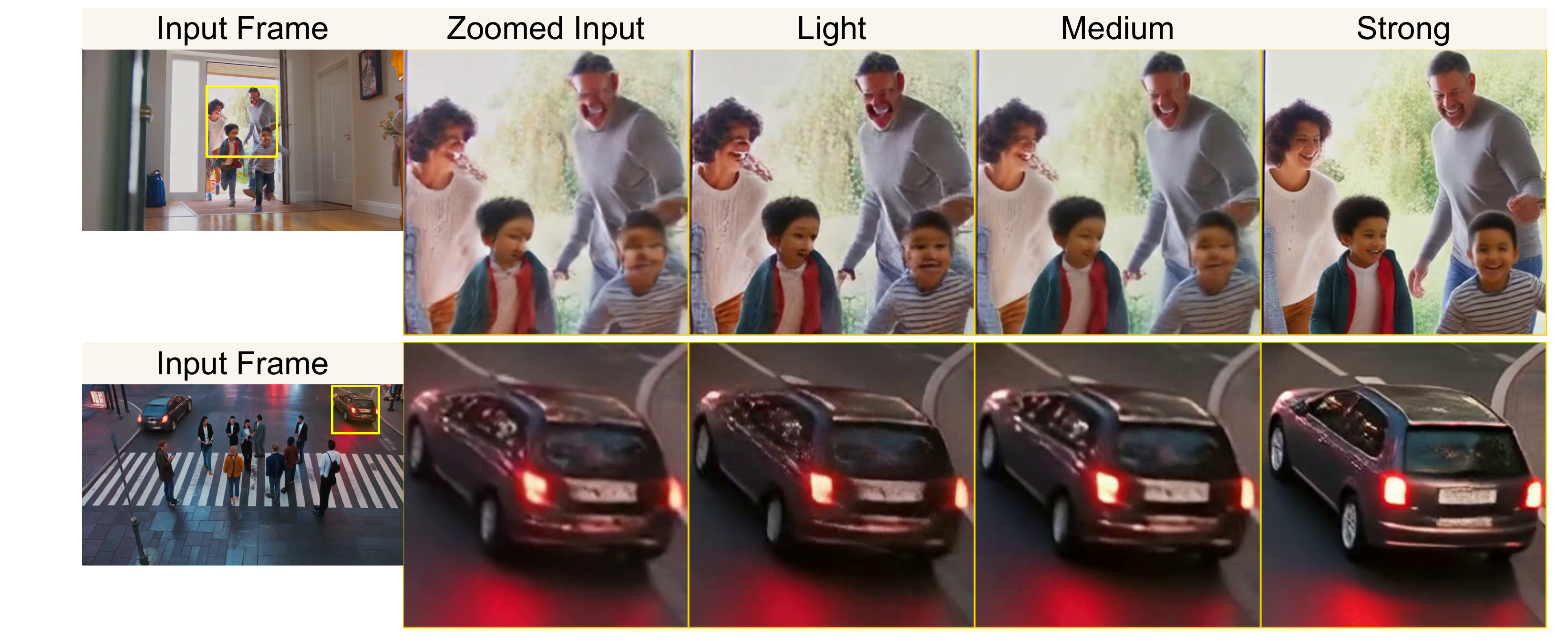}
    \vspace{-5mm}
   \caption{\textbf{Effect of degradation strength.} Stronger augmentations yield cleaner faces and sharper details.}
    \label{fig:strength_ablation}
\end{figure}

\noindent\textbf{Synthetic degradation strength during training.}
We ablate three augmentation levels—\emph{Light}, \emph{Medium}, and \emph{Strong}. All variants share the same backbone and training recipe; only the corruption schedule (blur, warping, morphing, frame dropping, temporal downsampling) is varied. As shown in Table.~\ref{tab:fiqa_strength}, \emph{Strong} yields the largest structure gains across all FIQA metrics on the AIGC54 benchmark, while \emph{Light} offers only mild sharpening and \emph{Medium} provides moderate improvements. GPT-based pairwise preferences (Table.~\ref{tab:preference_strength}) follow the same trend. Qualitatively (Fig.~\ref{fig:strength_ablation}), \emph{Strong} best removes geometric distortions without altering pose or identity.

Besides, Table.~\ref{tab:ref_benchmarks} indicates that on distortion-oriented SR/VR metrics (e.g., PSNR, SSIM on SPMCS and REDS30), \emph{Medium} slightly outperforms \emph{Strong}, likely because heavy synthetic degradations induce prior-driven corrections that deviate from pixel-wise ground truth. We therefore report both settings: \emph{Medium} for distortion benchmarks, and \emph{Strong} as the default for AIGC/real-world refinement.

\begin{figure}[t]
    \centering
    \includegraphics[width=\linewidth]{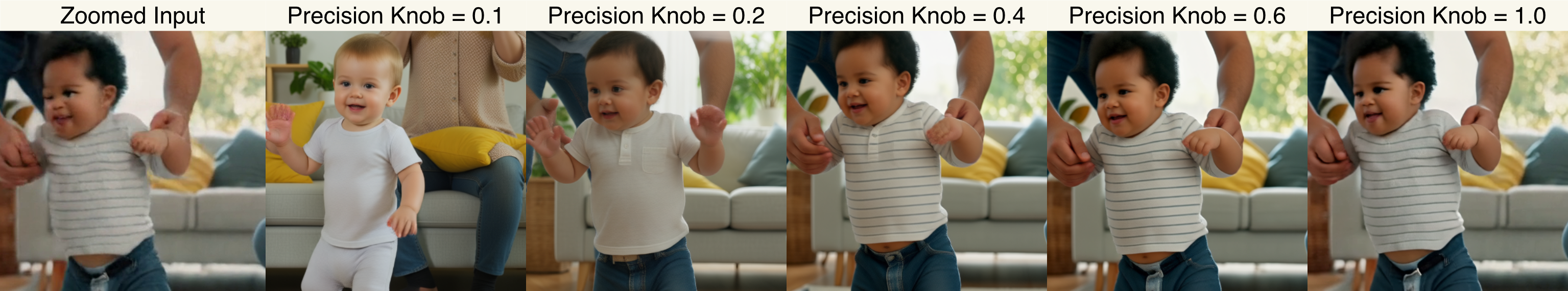}
    \vspace{-4mm}
   \caption{\textbf{Inference Precision knob.} High precision preserves input details; low precision enables stronger corrective synthesis.}
    \label{fig:creative_ablation}
\end{figure}

\noindent\textbf{Inference control scale.}
At test time we expose a single control scale $s$ that rescales all adapter gains, modulating how strongly the degraded video guides the frozen prior. As shown in Fig.~\ref{fig:creative_ablation}, higher $s$ values yield more precise, high-fidelity restoration close to the input, while lower $s$ allow more prior-driven, highly creative re-synthesis.

\noindent

\begin{table}[t]
\centering
\small
\begingroup
\setlength{\tabcolsep}{4pt}  %
\arrayrulecolor{gray!50}

\begin{tabular}{l|ccc}
\arrayrulecolor{black}
\toprule
\textbf{Quality Metric} & \multicolumn{3}{c}{\textbf{Degradation Strength}} \\
 & \textbf{Light} & \textbf{Medium} & \textbf{Strong} \\
\midrule
eDifFIQA  & \second{5.63} & 3.16 & \best{34.34} \\
DifFIQA   & 0.07          & \second{0.39} & \best{0.46} \\
CLIB-FIQA & 2.15          & \second{4.04} & \best{12.60} \\
CR-FIQA   & 6.01          & \second{7.03} & \best{15.95} \\
FaceQAN   & 5.02          & \second{9.45} & \best{26.14} \\
\bottomrule
\end{tabular}
\vspace{-3mm}
\caption{\textbf{FIQA gains vs.\ training strength.} Values are relative improvements (\%) over the input videos.}
\vspace{-1mm}
\label{tab:fiqa_strength}
\endgroup
\end{table}

\begin{table}[t]
\centering
\small
\begingroup
\setlength{\tabcolsep}{4pt}  %
\arrayrulecolor{gray!50}

\begin{tabular}{l|ccc}
\arrayrulecolor{black}
\toprule
\textbf{Scoring Aspect} & \multicolumn{3}{c}{\textbf{Degradation Strength}} \\
 & \textbf{Light} & \textbf{Medium} & \textbf{Strong} \\
\midrule
Visual Quality         & 7.69 & \second{7.89} & \best{8.20} \\
Temporal Consistency   & \second{8.52} & 8.46 & \best{8.77} \\
Face Quality           & \second{5.66} & 5.65 & \best{5.99} \\
Motion Realism         & \second{8.18} & 8.05 & \best{8.49} \\
Lighting \& Atmosphere & \second{8.60} & 8.56 & \best{8.84} \\
Detail Preservation    & 7.61 & \second{7.74} & \best{8.14} \\
Overall Mean           & 7.71 & \second{7.73} & \best{8.07} \\
\bottomrule
\end{tabular}
\vspace{-3mm}
\caption{Preference scores (1–10) vs.\ training strength}
\vspace{-3mm}
\label{tab:preference_strength}
\endgroup
\end{table}

\section{Conclusion}
We introduced CreativeVR, a unified diffusion-prior framework that refines both AIGC and real-world videos while preserving structure, motion, and details. By combining plug-and-play adapter modules with a temporally coherent corruption curriculum, CreativeVR corrects severe structural and temporal artifacts that traditional VR/VSR or post-hoc refiners fail to address. A controllable precision knob enables a smooth trade-off between fidelity-oriented restoration and prior-guided structural correction, making the method broadly applicable across diverse degradation regimes. Experiments on AIGC-artifact benchmarks, real degraded footage, and standard VR datasets show that CreativeVR achieves state-of-the-art restoration quality with practical throughput and strong zero-shot scalability. As a lightweight enhancement layer for frozen T2V backbones, CreativeVR provides a practical path toward production-ready video refinement in modern creative pipelines.

\clearpage
\appendix

\renewcommand{\thesection}{\Alph{section}}

\section*{Appendix Overview}
\begin{itemize}
    \item Implementation details, including dataset descriptions and evaluation metrics, are provided in~\Cref{suppsec:implementations}.
    \item Additional quantitative results on further benchmarks are reported in~\Cref{suppsec:quantitative}.
    \item Additional visualizations, qualitative comparisons, and diverse application examples are presented in~\Cref{suppsec:visualizations}.
\end{itemize}

\section{Datasets and Implementation Details}
\label{suppsec:implementations}
\subsection{Datasets}
For traditional video restoration, we \emph{only evaluate} on three standard synthetic benchmarks REDS30~\cite{nah2019ntire}, SPMCS~\cite{yi2019progressive}, and UDM10~\cite{tao2017detail} without any additional training or fine-tuning. Following prior work, we use their standard synthetic degradation pipelines and paired LR–HR sequences for evaluation, which cover diverse motion patterns and scene types and provide a controlled setting to measure both spatial fidelity and temporal consistency.
\subsection{Metrics}
For precision-oriented traditional video restoration with paired ground truth, we report four standard full-reference metrics: PSNR\cite{psnr} and SSIM\cite{ssim} to measure distortion fidelity, and LPIPS~\cite{lpips} and DISTS~\cite{dists} to capture perceptual similarity. All scores are averaged over all frames and sequences in each dataset.
\subsection{Synthetic Degradation pipeline}
\label{sec:synthetic_degradation}
We define three temporal augmentation presets - \textit{Light}, \textit{Medium},
and \textit{Strong}, that control the overall strength of temporal corruption.
For each clip, we first apply a single spatial augmentation and then compose a
small set of temporal operators in a fixed spatial-first ordering. On average,
Light uses roughly two temporal operators per clip, Medium uses three, and
Strong uses up to four, drawn from the pool described below.

\noindent\textbf{Motion blur.}
To simulate camera and object motion, we apply a 2D motion blur with a random
orientation in $[0^\circ, 360^\circ]$ and a kernel size sampled from a range on
the order of $3$--$20$ pixels. The blur strength is tied to the preset, with
Light sampling from the lower end of this range and Strong emphasizing longer
and more pronounced streaks.
\begin{figure}[t]
    \centering
    \includegraphics[width=\linewidth,trim=11 5 10 27,clip]{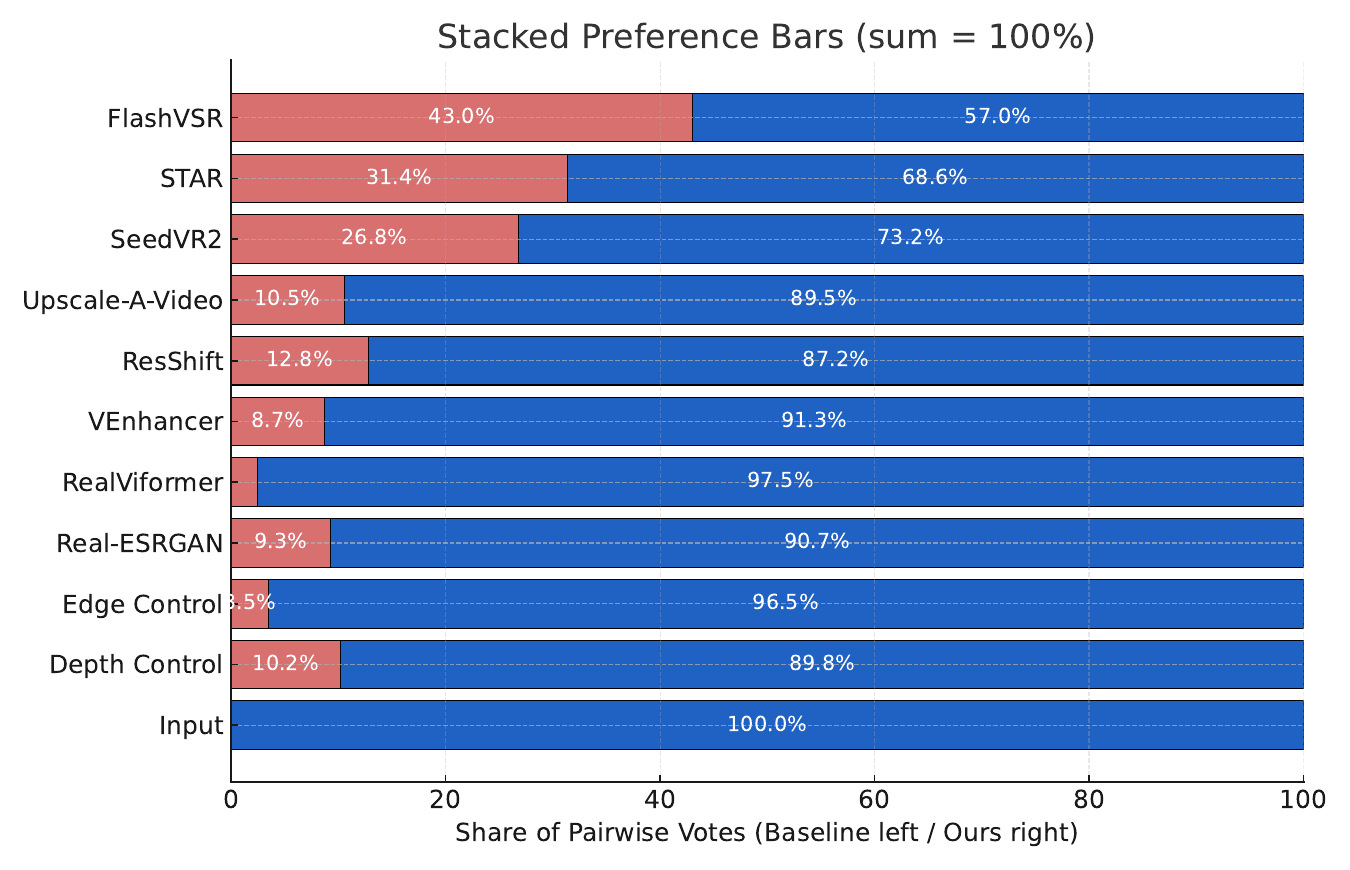}
    \caption{\textbf{Preference Study.} Bars show pairwise vote shares between \textit{Ours} (right) and each baseline (left); higher right-side bars indicate stronger preference for our method.}
    \label{fig:arena}
\end{figure}

\noindent\textbf{Warped grid distortion.}
Geometric wobble and mild rolling-shutter artifacts are introduced via a
grid-based spatial warp. We vary both the grid resolution (from coarse
$\sim\!4\times4$ to finer $\sim\!12\times12$ control points) and the
displacement amplitude (roughly $0.05$ to $0.3$ in normalized coordinates),
with Strong using coarser grids and larger displacements than Light.

\noindent\textbf{Temporal morphing.}
We synthesize nonlinear motion by interpolating between a small number of
keyframes (two in our implementation) using a parametric morphing operator.
A scalar strength parameter controls how far the interpolated motion deviates
from the original trajectory; Light uses mild perturbations, whereas Strong
samples from the upper part of the range (up to around $0.5$--$0.6$).

\begin{table*}[t]
\centering
\small
\begingroup
\setlength{\tabcolsep}{4pt}
\arrayrulecolor{gray!50}
\begin{tabular}{l|
        c|c|c|c|c|c|c|c|c}
\arrayrulecolor{black}
\toprule
\diagbox[
  width=2.8cm,
  height=1.2cm,
  trim=l,
  font=\scriptsize
]{\textbf{Performance}\\\textbf{Metric}}{\textbf{Restoration}\\\textbf{Method}}
& \begin{tabular}[c]{@{}c@{}}\textbf{FlashVSR}\\\cite{zhuang2025flashvsr}\end{tabular}
& \begin{tabular}[c]{@{}c@{}}\textbf{Real-}\\\textbf{ESRGAN}\\\cite{wang2021real}\end{tabular}
& \begin{tabular}[c]{@{}c@{}}\textbf{Real-}\\\textbf{Viformer}\\\cite{zhang2024realviformer}\end{tabular}
& \begin{tabular}[c]{@{}c@{}}\textbf{ResShift}\\\cite{yue2023resshift}\end{tabular}
& \begin{tabular}[c]{@{}c@{}}\textbf{SeedVR2}\\\cite{wang2025seedvr2}\end{tabular}
& \begin{tabular}[c]{@{}c@{}}\textbf{STAR}\\\cite{xie2025star}\end{tabular}
& \begin{tabular}[c]{@{}c@{}}\textbf{Upscale-A}\\\textbf{-Video}\\\cite{xu2024upscaleavideo}\end{tabular}
& \begin{tabular}[c]{@{}c@{}}\textbf{VEnhancer}\\\cite{he2024venhancer}\end{tabular}
& \textbf{Ours} \\
\midrule
Aesthetic Quality        & 0.6313 & 0.6338 & 0.6334 & 0.6284 & \best{0.6519} & 0.6157 & 0.6085 & 0.6314 & \second{0.6340} \\
Background Consistency   & 0.9486 & 0.9481 & 0.9477 & 0.9503 & 0.9416 & \second{0.9506} & 0.9493 & \best{0.9565} & 0.9493 \\
Imaging Quality          & 0.7229 & 0.7201 & 0.7075 & 0.6927 & 0.6963 & \best{0.7347} & 0.7078 & 0.6999 & \second{0.7311} \\
Motion Smoothness        & 0.9841 & 0.9804 & 0.9853 & 0.9843 & 0.9850 & 0.9827 & 0.9856 & \best{0.9907} & \second{0.9879} \\
Dynamic Degree           & \second{0.9118} & 0.8824 & 0.8761 & 0.8655 & 0.8529 & 0.8800 & 0.8529 & 0.8824 & \best{0.9444} \\
Subject Consistency      & 0.9381 & 0.9385 & 0.9384 & 0.9384 & 0.9343 & 0.9363 & 0.9375 & \best{0.9413} & \second{0.9387} \\
\hline
Average                  & \second{0.8561} & 0.8506 & 0.8481 & 0.8433 & 0.8437 & 0.8500 & 0.8403 & 0.8504 & \best{0.8642} \\
\bottomrule
\end{tabular}
\arrayrulecolor{black}
\endgroup
\caption{\textbf{VBench perceptual scores.} Higher is better for all metrics. Our method achieves the best average score across all six dimensions, with especially strong gains in dynamic degree and imaging quality while remaining competitive on subject/background consistency, aesthetic quality, and motion smoothness.}
\label{tab:vbench_scores}
\end{table*}

\noindent\textbf{Stochastic frame dropping.}
Dropped or corrupted frames are simulated by randomly removing frames and
reconstructing them via simple temporal interpolation. We control both the
per-frame drop probability and the maximum number of dropped frames per clip,
with Light using low drop rates (single-frame events) and Strong allowing
multiple consecutive drops, leading to visibly irregular motion.

\noindent\textbf{Temporal downsampling.}
Finally, we degrade temporal fidelity by subsampling the clip in time with a
factor between roughly $1.5$ and $3.5$, followed by re-timing to the original
length. Higher factors are reserved for the Strong preset and produce
noticeable temporal aliasing and “jerky” motion, while Light uses only mild
subsampling.

\noindent\textbf{Selection strategy.}
Given a preset, we randomly sample a subset of the above operators and apply
them sequentially, ensuring that Light, Medium, and Strong correspond to
increasing numbers and strengths of temporal corruptions while keeping the
underlying action recognizable for training.

\section{Additional Quantitative Results}
\label{suppsec:quantitative}
\subsection{Pairwise Preference Evaluation (Arena)}
We further employ a pairwise arena-style evaluation using a GPT judge. For each comparison, the evaluator is shown two restored clips from the same input sequence and asked to indicate a preference based on temporal consistency and frame quality. The aggregated votes across all clips provide a direct perceptual ranking. As shown in Fig.~\ref{fig:arena}, our method receives the majority of votes against both classical and diffusion-based baselines, reflecting strong perceptual preference in head-to-head comparisons. We also conducted a human check on randomly sampled pairs, using 10 pairs per head-to-head comparison($\approx 20\%$ of all comparisons), where four independent users confirmed that GPT's pairwise preferences were consistent with majority human choices, although the arena contributed limited additional insights beyond the multi-aspect scores.
We also ran a small human check on 20\% of the arena pairs, confirming that GPT's preferences aligned with majority human choices, although the arena added limited additional insights beyond the multi-aspect scores.

\subsection{Evaluation on VBench Benchmark}
We evaluate perceptual video quality on VBench~\cite{huang2024vbench}, a comprehensive benchmark that decomposes ``video generation quality'' into multiple human-aligned evaluation dimensions with dedicated automatic metrics for each. In \Cref{tab:vbench_scores}, we report the temporal-quality dimensions \textit{subject\_consistency}, \textit{background\_consistency}, \textit{motion\_smoothness}, and \textit{dynamic\_degree}, together with frame-wise \textit{aesthetic\_quality} and \textit{imaging\_quality} for generations on our \textit{AIGC54} set. Briefly, subject and background consistency measure how well the appearance of the main subject and the background remain coherent across frames, \textit{motion\_smoothness} quantifies whether motions follow smooth and physically plausible trajectories, dynamic degree captures the amount of motion present in the video, and aesthetic and imaging quality assess the visual appeal and low-level distortions (e.g., blur, noise, over-exposure) of individual frames.

As shown in \Cref{tab:vbench_scores}, our method achieves the highest overall VBench average score (0.8642), outperforming all competing video restoration baselines, including FlashVSR (0.8561). It obtains the best \textit{dynamic\_degree} (0.9444), indicating that our model can generate videos with stronger and more realistic motion while still maintaining temporal coherence. At the same time, our approach remains highly competitive on \textit{subject\_consistency} (0.9387) and \textit{background\_consistency} (0.9493), and reaches the second-best performance on \textit{imaging\_quality} (0.7311) and \textit{motion\_smoothness} (0.9879), suggesting that the recovered videos are both sharp and temporally smooth. Finally, our \textit{aesthetic\_quality} score (0.6340) is comparable to or better than most baselines, showing that the refined videos are visually pleasing while improving dynamic realism.
\begin{figure*}[!t]
  \centering
  \begin{minipage}{0.88\textwidth}
\begin{lstlisting}[language=Python]
# Qwen2.5-VL prompt template for object-centric quality
JUDGE_PROMPT = """You are a meticulous computer-vision rater for a CVPR paper.
Judge only the cropped object/person region. Be strict about geometric
plausibility, boundary stability, and structural artifacts (wobble, double
edges, ghosting, halos, smeared surfaces, plastic textures, ringing). If an
'INPUT' image or caption is provided, **ignore it** completely -- judge each
candidate independently. Return ONLY valid JSON. No extra text. If a criterion
is not visible, set it to null and add a brief tag in "notes".

TASK
Score each candidate crop A...K using 0-10 integers (0 = very poor,
10 = excellent). Use labels exactly as given (A, B, C, ...).

Rubric (0-10, integers):
- GE (Geometry & Silhouette): Judge geometry in the object's semantic frame.
  Check canonical parts/primitives for the category (e.g., limb/joint
  alignment, straight planar borders, right-angle corners, circular rims/wheels,
  text baselines, symmetry where applicable). Penalize distorted contours,
  wobble/jitter, double edges, part warping, melted forms, bent straight lines,
  ovalized circles, or misshapen letters/logos.
- ED (Edge Definition & Continuity): Assess edges along semantic boundaries
  (main silhouette, structural lines, text/glyph strokes, circular/straight
  borders). Edges must trace the true shape with continuous, coherent paths.
  Do not reward edge contrast by itself. Penalize fraying, ghost/double edges,
  bright/dark halos/ringing, staircase/zippering, or overshoot that shifts the
  perceived boundary.
- DT (Detail Realism): Reward fine detail only when it follows the
  form/material (fabric folds conforming to shape, hair/filament strands,
  micro-structure, legible glyph strokes). Penalize plasticiness,
  over-smoothing, and especially invented crunchy micro-patterns or grain that
  do not align with the material or introduce false pores/texture.
- AR (Artifact Level; reverse scale): Structural artifacts such as
  halos/ringing/overshoot, checkerboard/tiling, zippering, deconvolution
  ripples, temporal smears, motion trails, compression-like textures,
  patch seams, moire. (10 = no artifacts, 0 = severe artifacts)
- TN (Tonal Robustness): Consider only major tonal failures -- blown highlights,
  crushed shadows, or hue shifts that harm form perception. Ignore
  vibrance/saturation; do not reward colorfulness.

Scoring rules:
- Integers only (0-10). When uncertain, choose the lower score.
- Prioritize shape correctness over sharpness. If you observe
  halo/overshoot/double contours or invented high-frequency texture, give low
  ED and AR (0-2) and do not boost DT for "crispness".
- Focus on structure and boundary integrity; ignore style/brightness preferences
  or artistic color.
- Heavily penalize geometry wobble, melted forms, halos/ghosting, and
  broken/duplicated edges.
- If a criterion cannot be judged, set it to null and add a tag in "notes"
  (e.g., "NOT_VISIBLE", "OCCLUDED").

OUTPUT JSON SCHEMA (strict):
{
  "per_image": {
    "A": {"GE": int|null, "ED": int|null, "DT": int|null,
          "AR": int|null, "TN": int|null, "notes": ["TAG1","TAG2"]},
    "B": {...}
  },
  "overall_ranking": ["C","A","B", ...],
  "confidence": "high" | "medium" | "low"
}

IMAGES:
{image_descriptions}
"""
\end{lstlisting}
  \end{minipage}
  \vspace{-4mm}
  \caption{Qwen2.5-VL prompt template for multi-aspect face-quality (rubric) scoring.}
  \label{rubric:qwen_rubric}
\end{figure*}

\begin{table*}[t]
\centering
\small
\begingroup
\setlength{\tabcolsep}{3pt}
\arrayrulecolor{gray!50}
\resizebox{\linewidth}{!}{%
\begin{tabular}{l|
                c|c|c|c|c|c|c|c|c}
\arrayrulecolor{black}
\toprule
\diagbox[
  width=2.9cm,
  height=1.4cm,
  trim=l,
  font=\scriptsize
]{\textbf{Performance}\\\textbf{Metric}}{\textbf{Restoration}\\\textbf{Method}}
& \begin{tabular}[c]{@{}c@{}}\textbf{FlashVSR}\\\cite{zhuang2025flashvsr}\end{tabular}
& \begin{tabular}[c]{@{}c@{}}\textbf{Real-}\\\textbf{ESRGAN}\\\cite{wang2021real}\end{tabular}
& \begin{tabular}[c]{@{}c@{}}\textbf{Real-}\\\textbf{Viformer}\\\cite{zhang2024realviformer}\end{tabular}
& \begin{tabular}[c]{@{}c@{}}\textbf{ResShift}\\\cite{yue2023resshift}\end{tabular}
& \begin{tabular}[c]{@{}c@{}}\textbf{SeedVR2}\\\cite{wang2025seedvr2}\end{tabular}
& \begin{tabular}[c]{@{}c@{}}\textbf{STAR}\\\cite{xie2025star}\end{tabular}
& \begin{tabular}[c]{@{}c@{}}\textbf{Upscale-A}\\\textbf{-Video}\\\cite{xu2024upscaleavideo}\end{tabular}
& \begin{tabular}[c]{@{}c@{}}\textbf{VEnhancer}\\\cite{he2024venhancer}\end{tabular}
& \textbf{Ours} \\
\midrule
Geometry \& Silhouette        & \second{7.64} & 5.98 & 4.50 & 2.46 & 4.28 & 5.20 & 3.82 & 6.48 & \best{8.04} \\
Edge Definition \& Continuity & \second{8.10} & 6.20 & 5.08 & 2.84 & 4.50 & 5.42 & 4.08 & 6.64 & \best{8.24} \\
Detail Realism                & \best{8.22}   & 5.54 & 4.14 & 2.34 & 3.90 & 4.80 & 3.58 & 6.02 & \second{7.46} \\
Artifact Level (reverse)      & 5.99         & 5.64 & 4.18 & 2.34 & 4.00 & 4.92 & 3.64 & \second{6.18} & \best{7.64} \\
Tonal Robustness              & \second{6.62} & 5.90 & 4.26 & 2.40 & 4.10 & 5.00 & 3.68 & 6.26 & \best{7.80} \\
\hline
Aggregate Score               & \second{7.31} & 5.85 & 4.43 & 2.48 & 4.16 & 5.07 & 3.76 & 6.32 & \best{7.84} \\
\bottomrule
\end{tabular}}
\arrayrulecolor{black}
\endgroup
\caption{\textbf{Multi-Aspect Scoring for Face Quality judged by Qwen2.5-VL-72B.} Scores (0--10) are given for geometry \& silhouette, edge definition \& continuity, detail realism, artifact level (reverse), and tonal robustness. Our method achieves the best aggregate score and leads on four out of five dimensions, corroborating the strong facial quality observed with FaceIQA.}
\label{tab:qwen_ranking}
\end{table*}

\begin{table*}[t]
\centering
\small
\begingroup
\setlength{\tabcolsep}{3pt}
\arrayrulecolor{gray!50}
\resizebox{\linewidth}{!}{%
\begin{tabular}{l| l|
                c| c| c| c| c| c| c| c| c| c}
\arrayrulecolor{black}
\toprule
\textbf{Corruption Type} & \textbf{Metric}
& \begin{tabular}[c]{@{}c@{}}\textbf{FlashVSR}\\\cite{zhuang2025flashvsr}\end{tabular}
& \begin{tabular}[c]{@{}c@{}}\textbf{Real-}\\\textbf{ESRGAN}\\\cite{wang2021real}\end{tabular}
& \begin{tabular}[c]{@{}c@{}}\textbf{Real-}\\\textbf{Viformer}\\\cite{zhang2024realviformer}\end{tabular}
& \begin{tabular}[c]{@{}c@{}}\textbf{ResShift}\\\cite{yue2023resshift}\end{tabular}
& \begin{tabular}[c]{@{}c@{}}\textbf{SeedVR2}\\\cite{wang2025seedvr2}\end{tabular}
& \begin{tabular}[c]{@{}c@{}}\textbf{STAR}\\\cite{xie2025star}\end{tabular}
& \begin{tabular}[c]{@{}c@{}}\textbf{Upscale-A}\\\textbf{-Video}\\\cite{xu2024upscaleavideo}\end{tabular}
& \begin{tabular}[c]{@{}c@{}}\textbf{VEnhancer}\\\cite{he2024venhancer}\end{tabular}
& \begin{tabular}[c]{@{}c@{}}\textbf{Ours}\\\textbf{(Medium)}\end{tabular}
& \begin{tabular}[c]{@{}c@{}}\textbf{Ours}\\\textbf{(Strong)}\end{tabular} \\
\midrule

\multirow{8}{*}{\begin{tabular}[c]{@{}l@{}}\textbf{Spatial}\\ Downsampling\end{tabular}}
& PSNR $\uparrow$     & 25.87 & 25.95 & \second{26.03} & 25.73 & 22.59 & 24.33 & 24.03 & 15.46 & \best{27.12} & 26.02 \\
& SSIM $\uparrow$     & 0.73  & 0.72  & 0.72          & 0.73  & 0.65  & 0.70  & 0.64  & 0.44  & \best{0.79} & \second{0.76} \\
& LPIPS $\downarrow$  & 0.38  & 0.40  & 0.43          & 0.18  & 0.36  & 0.21  & 0.28  & 0.38  & \best{0.13} & \second{0.16} \\
& DISTS $\downarrow$  & 0.15  & 0.18  & 0.17          & \second{0.08} & 0.13  & \second{0.08} & 0.12  & 0.10  & \best{0.06} & \best{0.06} \\
& NIQE $\downarrow$   & \second{2.91} & 6.63  & 6.51  & 3.09  & 5.94  & 3.39  & \best{2.49} & 4.20  & 3.00 & 3.17 \\
& MUSIQ $\uparrow$    & 66.40 & 36.48 & \best{67.39}   & 27.11 & 31.62 & 64.45 & \second{66.79} & 48.33 & 62.81 & 62.98 \\
& CLIP-IQA $\uparrow$ & \second{0.53} & 0.37  & \best{0.54}   & 0.33  & 0.35  & 0.48  & 0.50  & 0.40  & 0.46 & 0.44 \\
& DOVER $\uparrow$    & 31.97 & 21.85 & \second{33.67} & 14.99 & 17.80 & \best{35.83} & 24.41 & 24.37 & 33.28 & 32.39 \\
\midrule

\multirow{8}{*}{\begin{tabular}[c]{@{}l@{}}\textbf{Spatio-Temporal}\\Downsampling \end{tabular}}
& PSNR $\uparrow$     & 15.48 & 25.13 & \best{25.86}   & 24.57 & 25.43 & 24.07 & 23.41 & 22.40 & 25.47 & \second{25.51} \\
& SSIM $\uparrow$     & 0.44  & 0.69  & 0.71          & 0.69  & \second{0.72} & 0.69  & 0.62  & 0.63  & \best{0.74} & \best{0.74} \\
& LPIPS $\downarrow$  & 0.39  & 0.41  & 0.43          & 0.21  & 0.38  & 0.23  & 0.30  & 0.39  & \best{0.16} & \second{0.18} \\
& DISTS $\downarrow$  & 0.10  & 0.19  & 0.17          & \second{0.09} & 0.14  & \second{0.09} & 0.13  & 0.15  & \best{0.07} & \best{0.07} \\
& NIQE $\downarrow$   & \second{2.95} & 6.58  & 6.52  & 3.09  & 5.83  & 3.43  & \best{2.53} & 4.34  & 2.97 & 3.21 \\
& MUSIQ $\uparrow$    & \second{66.34} & 34.02 & 26.60 & 66.13 & 33.01 & 60.65 & \best{66.57} & 43.92 & 62.56 & 61.71 \\
& CLIP-IQA $\uparrow$ & \second{0.53} & 0.37  & \best{0.54}   & 0.34  & 0.36  & 0.46  & 0.50  & 0.39  & 0.46 & 0.44 \\
& DOVER $\uparrow$    & 32.36 & 22.09 & \second{36.98} & 14.94 & 18.39 & 33.64 & 24.10 & 23.19 & \best{37.02} & 31.56 \\
\midrule

\multirow{8}{*}{\begin{tabular}[c]{@{}l@{}}\textbf{Spatio-Temporal}\\Light\end{tabular}}
& PSNR $\uparrow$     & 20.73 & \second{20.84} & 20.65 & 20.24 & 20.59 & 20.56 & 19.91 & 19.92 & 20.73 & \best{21.18} \\
& SSIM $\uparrow$     & \second{0.58} & \best{0.59} & \second{0.58} & 0.56 & \best{0.59} & \best{0.59} & 0.54  & 0.56  & \second{0.58} & \best{0.59} \\
& LPIPS $\downarrow$  & 0.64  & 0.59  & \best{0.37}   & 0.46  & 0.48  & 0.54  & 0.47  & 0.66  & \second{0.43} & 0.51 \\
& DISTS $\downarrow$  & 0.34  & 0.32  & \second{0.20} & 0.25  & 0.26  & 0.29  & 0.25  & 0.36  & \best{0.13} & 0.24 \\
& NIQE $\downarrow$   & 8.23  & 8.13  & 4.03          & 4.44  & 5.96  & 6.85  & \second{3.88} & 8.61  & \best{3.20} & 6.62 \\
& MUSIQ $\uparrow$    & 15.64 & 22.04 & 46.37         & 44.34 & 23.97 & 21.88 & \second{51.76} & 21.47 & 63.89 & 21.34 \\
& CLIP-IQA $\uparrow$ & 0.24  & 0.31  & 0.40          & \second{0.49} & 0.28  & 0.29  & 0.43  & 0.30  & \best{0.51} & 0.30 \\
& DOVER $\uparrow$    & 5.86  & 11.91 & \second{22.05} & 20.30 & 12.96 & 13.61 & 17.54 & 9.75  & \best{27.99} & 10.53 \\
\midrule

\multirow{8}{*}{\begin{tabular}[c]{@{}l@{}}\textbf{Spatio-Temporal}\\Strong\end{tabular}}
& PSNR $\uparrow$     & 20.59 & 15.63 & 20.48 & 20.12 & 20.41 & 20.42 & 19.73 & 19.75 & \second{20.69} & \best{20.95} \\
& SSIM $\uparrow$     & \second{0.58} & 0.43  & \second{0.58} & 0.56 & \second{0.58} & \second{0.58} & 0.53 & 0.56 & \second{0.58} & \best{0.59} \\
& LPIPS $\downarrow$  & 0.64  & 0.59  & \best{0.38}   & 0.51  & 0.49  & 0.55  & 0.48  & 0.66  & 0.47 & \second{0.43} \\
& DISTS $\downarrow$  & 0.34  & 0.32  & \second{0.20} & 0.26  & 0.26  & 0.29  & 0.25  & 0.36  & 0.24 & \best{0.13} \\
& NIQE $\downarrow$   & 8.28  & 8.12  & 3.89          & 4.47  & 5.92  & 6.84  & \second{3.74} & 8.59  & 6.56 & \best{3.15} \\
& MUSIQ $\uparrow$    & 15.75 & 21.74 & 47.60         & 43.47 & 23.94 & 21.83 & \second{53.17} & 21.50 & 21.95 & \best{63.19} \\
& CLIP-IQA $\uparrow$ & 0.24  & 0.31  & 0.41          & \second{0.49} & 0.28  & 0.30  & 0.43  & 0.29  & 0.30 & \best{0.51} \\
& DOVER $\uparrow$    & 5.48  & 11.11 & \second{22.50} & 19.51 & 12.83 & 13.54 & 17.76 & 9.25  & 10.48 & \best{28.39} \\
\bottomrule
\end{tabular}}
\arrayrulecolor{black}
\endgroup
\caption{\textbf{Robustness under synthetic corruptions on REDS30~\cite{nah2019ntire}.} Higher is better for PSNR, SSIM, MUSIQ, CLIP-IQA, and DOVER; lower is better for LPIPS, DISTS, and NIQE.}
\label{tab:corruption_reds30}
\end{table*}

\begin{table*}[t]
\centering
\small
\begingroup
\setlength{\tabcolsep}{3pt}
\arrayrulecolor{gray!50}
\resizebox{\linewidth}{!}{%
\begin{tabular}{l| l|
                c| c| c| c| c| c| c| c| c| c}
\arrayrulecolor{black}
\toprule
\textbf{Augmentation Type} & \textbf{Metric}
& \begin{tabular}[c]{@{}c@{}}\textbf{FlashVSR}\\\cite{zhuang2025flashvsr}\end{tabular}
& \begin{tabular}[c]{@{}c@{}}\textbf{Real-}\\\textbf{ESRGAN}\\\cite{wang2021real}\end{tabular}
& \begin{tabular}[c]{@{}c@{}}\textbf{Real-}\\\textbf{Viformer}\\\cite{zhang2024realviformer}\end{tabular}
& \begin{tabular}[c]{@{}c@{}}\textbf{ResShift}\\\cite{yue2023resshift}\end{tabular}
& \begin{tabular}[c]{@{}c@{}}\textbf{SeedVR2}\\\cite{wang2025seedvr2}\end{tabular}
& \begin{tabular}[c]{@{}c@{}}\textbf{STAR}\\\cite{xie2025star}\end{tabular}
& \begin{tabular}[c]{@{}c@{}}\textbf{Upscale-A}\\\textbf{-Video}\\\cite{xu2024upscaleavideo}\end{tabular}
& \begin{tabular}[c]{@{}c@{}}\textbf{VEnhancer}\\\cite{he2024venhancer}\end{tabular}
& \begin{tabular}[c]{@{}c@{}}\textbf{Ours}\\\textbf{(Medium)}\end{tabular}
& \begin{tabular}[c]{@{}c@{}}\textbf{Ours}\\\textbf{(Strong)}\end{tabular} \\
\midrule

\multirow{8}{*}{\begin{tabular}[c]{@{}l@{}}\textbf{Spatial}\\Downsampling\end{tabular}}
& PSNR $\uparrow$     & 25.6  & \second{26.4} & 25.62 & 25.98 & 24.2  & 16.06 & 24.45 & 22.18 & \best{27.2}  & 26.31 \\
& SSIM $\uparrow$     & 0.722 & 0.715        & \second{0.727} & 0.713 & 0.702 & 0.417 & 0.633 & 0.63  & \best{0.767} & 0.726 \\
& LPIPS $\downarrow$  & \second{0.193} & 0.367 & 0.31  & 0.203 & 0.378 & 0.274 & 0.285 & 0.402 & \best{0.182} & 0.354 \\
& DISTS $\downarrow$  & \best{0.09} & 0.189 & 0.137 & \second{0.1} & 0.131 & 0.115 & 0.133 & 0.169 & 0.109 & 0.154 \\
& NIQE $\downarrow$   & \second{3.29} & 6.99  & 6.18  & 3.6   & 3.84  & 4.91  & \best{3.04} & 5.58 & 3.76 & 6.89 \\
& MUSIQ $\uparrow$    & \second{68.73} & 43.28 & 42.75 & \best{70.42} & 64.7  & 56    & 67.92 & 41.62 & 66.54 & 35.15 \\
& CLIP-IQA $\uparrow$ & \second{0.568} & 0.396 & 0.412 & \best{0.609} & 0.513 & 0.425 & 0.506 & 0.406 & 0.512 & 0.366 \\
& DOVER $\uparrow$    & 51.12 & 38.64 & 36.12 & \second{53.39} & 53.03 & 50.24 & 43.22 & 40.92 & \best{53.92} & 32.17 \\
\midrule

\multirow{8}{*}{\begin{tabular}[c]{@{}l@{}}\textbf{Spatio-Temporal}\\Downsampling\end{tabular}}
& PSNR $\uparrow$     & 16.19 & 25.23 & 25.51 & 24.67 & 24.77 & 23.59 & 23.57 & 21.85 & \best{25.84} & \second{24.64} \\
& SSIM $\uparrow$     & 0.418 & 0.689 & \best{0.717} & 0.709 & 0.7   & 0.684 & 0.612 & 0.619 & \second{0.715} & 0.674 \\
& LPIPS $\downarrow$  & 0.387 & 0.385 & 0.215 & 0.304 & \second{0.209} & 0.293 & 0.307 & 0.43  & 0.359 & \best{0.236} \\
& DISTS $\downarrow$  & 0.136 & 0.194 & 0.121 & 0.133 & 0.095 & 0.126 & 0.143 & 0.191 & 0.156 & \best{0.114} \\
& NIQE $\downarrow$   & 3.34  & 6.98  & 3.75  & 5.94  & 3.91  & 4.96  & \best{3.04} & 5.77 & 6.89 & \second{3.68} \\
& MUSIQ $\uparrow$    & \second{69.01} & 41.06 & 64.51 & 45.77 & 63.28 & 52.74 & 67.54 & 38.15 & 34.97 & \best{69.07} \\
& CLIP-IQA $\uparrow$ & 0.572 & 0.394 & 0.507 & 0.421 & 0.507 & 0.413 & 0.51  & 0.401 & 0.369 & \best{0.605} \\
& DOVER $\uparrow$    & 52.05 & 38.53 & \second{53.65} & 37.26 & 53.1  & 47.68 & 42.99 & 40.13 & 32.44 & \best{53.26} \\
\midrule

\multirow{8}{*}{\begin{tabular}[c]{@{}l@{}}\textbf{Spatio-Temporal}\\Light\end{tabular}}
& PSNR $\uparrow$     & 21.56 & \best{21.9} & 21.65 & 21.21 & 21.47 & 21.17 & 21.27 & 16.49 & 20.43 & \second{21.9} \\
& SSIM $\uparrow$     & 0.595 & 0.602 & 0.602 & 0.573 & 0.597 & 0.6   & 0.572 & 0.426 & 0.576 & \best{0.608} \\
& LPIPS $\downarrow$  & 0.547 & 0.53  & 0.388 & 0.445 & 0.437 & 0.543 & 0.474 & 0.602 & \second{0.407} & \best{0.432} \\
& DISTS $\downarrow$  & 0.298 & 0.302 & 0.217 & 0.26  & 0.242 & 0.305 & 0.27  & 0.345 & \best{0.149} & \second{0.214} \\
& NIQE $\downarrow$   & 8.36  & 8.4   & 4.76  & 4.95  & 6.31  & 7.69  & 5.41  & 8.74 & \best{3.62} & \second{6.47} \\
& MUSIQ $\uparrow$    & 17.32 & 24.42 & 45.66 & \second{47.71} & 27.53 & 24.87 & 42.74 & 23.47 & \best{67.23} & 30.24 \\
& CLIP-IQA $\uparrow$ & 0.278 & 0.349 & 0.458 & \best{0.569} & 0.32  & 0.31  & 0.404 & 0.336 & \second{0.55} & 0.37 \\
& DOVER $\uparrow$    & 22.27 & 30.22 & 43.64 & 43.63 & 36.11 & 33.97 & 31.95 & 29.67 & \best{49.22} & \second{30.02} \\
\midrule

\multirow{8}{*}{\begin{tabular}[c]{@{}l@{}}\textbf{Spatio-Temporal}\\Strong\end{tabular}}
& PSNR $\uparrow$     & 21.69 & \best{22.04} & 21.75 & 21.37 & 16.46 & 21.31 & 21.29 & 20.55 & 21.73 & \second{21.92} \\
& SSIM $\uparrow$     & 0.606 & \second{0.608} & 0.607 & 0.581 & 0.431 & 0.606 & 0.571 & 0.581 & 0.601 & \best{0.611} \\
& LPIPS $\downarrow$  & 0.426 & 0.521 & \best{0.371} & 0.431 & 0.54  & 0.527 & 0.446 & 0.587 & \second{0.401} & 0.421 \\
& DISTS $\downarrow$  & 0.238 & 0.298 & \second{0.208} & 0.25  & 0.293 & 0.294 & 0.248 & 0.33  & \best{0.147} & 0.209 \\
& NIQE $\downarrow$   & 6.22  & 8.35  & \second{4.61} & 4.85  & 8.32  & 7.48  & 4.81  & 8.48 & \best{3.55} & 6.27 \\
& MUSIQ $\uparrow$    & 28.88 & 24.87 & 48.3  & \second{48.39} & 23.94 & 26    & 47.25 & 23.65 & \best{66.84} & 33.77 \\
& CLIP-IQA $\uparrow$ & 0.331 & 0.35  & 0.469 & \best{0.569} & 0.282 & 0.318 & 0.431 & 0.33  & \second{0.546} & 0.386 \\
& DOVER $\uparrow$    & 32.94 & 29.03 & 40.45 & \second{40.94} & 32.25 & 20.59 & 27.11 & 28.47 & \best{47.35} & 27.93 \\
\bottomrule
\end{tabular}}
\arrayrulecolor{black}
\endgroup
\caption{\textbf{Robustness under learned augmentation strengths on YouHQ40~\cite{xu2024upscaleavideo}}. Higher is better for PSNR, SSIM, MUSIQ, CLIP-IQA, and DOVER; lower is better for LPIPS, DISTS, and NIQE.}
\label{tab:corruption_youhq}
\end{table*}

\subsection{Multi Aspect Scoring Face Quality}
Beyond FaceIQA, which already shows that our method significantly improves
facial quality on AIGC54, we further run an object-centric evaluation in which
Qwen2.5-VL-72B acts as a geometry-aware judge on cropped face regions.
For each crop, Qwen scores five aspects following our prompt rubric:
\emph{Geometry \& Silhouette} (GE), \emph{Edge Definition \& Continuity} (ED),
\emph{Detail Realism} (DT), \emph{Artifact Level} (AR; reverse scale), and
\emph{Tonal Robustness} (TN). The model outputs integer scores in $[0,10]$,
and we average them across frames and dimensions to obtain an aggregate
object-centric quality score. The full prompt and scoring rubric are provided
in the rubric shown in ~\Cref{rubric:qwen_rubric}.

As summarized in \Cref{tab:qwen_ranking}, our method achieves the highest
aggregate score (7.84), providing an independent confirmation of the strong
facial quality observed with FaceIQA. Compared to prior video restoration
baselines, we obtain the best scores on \emph{Geometry \& Silhouette},
\emph{Edge Definition \& Continuity}, \emph{Artifact Level}, and \emph{Tonal Robustness}, indicating more stable
facial shapes, cleaner boundaries, and fewer structural artifacts while
preserving photometric fidelity. Although FlashVSR attains slightly higher
\emph{Detail Realism}, our method remains competitive and does so without
introducing spurious high-frequency artifacts, as reflected by our superior
geometry and artifact scores.

\begin{figure*}[t]
    \centering
    \includegraphics[width=\linewidth]{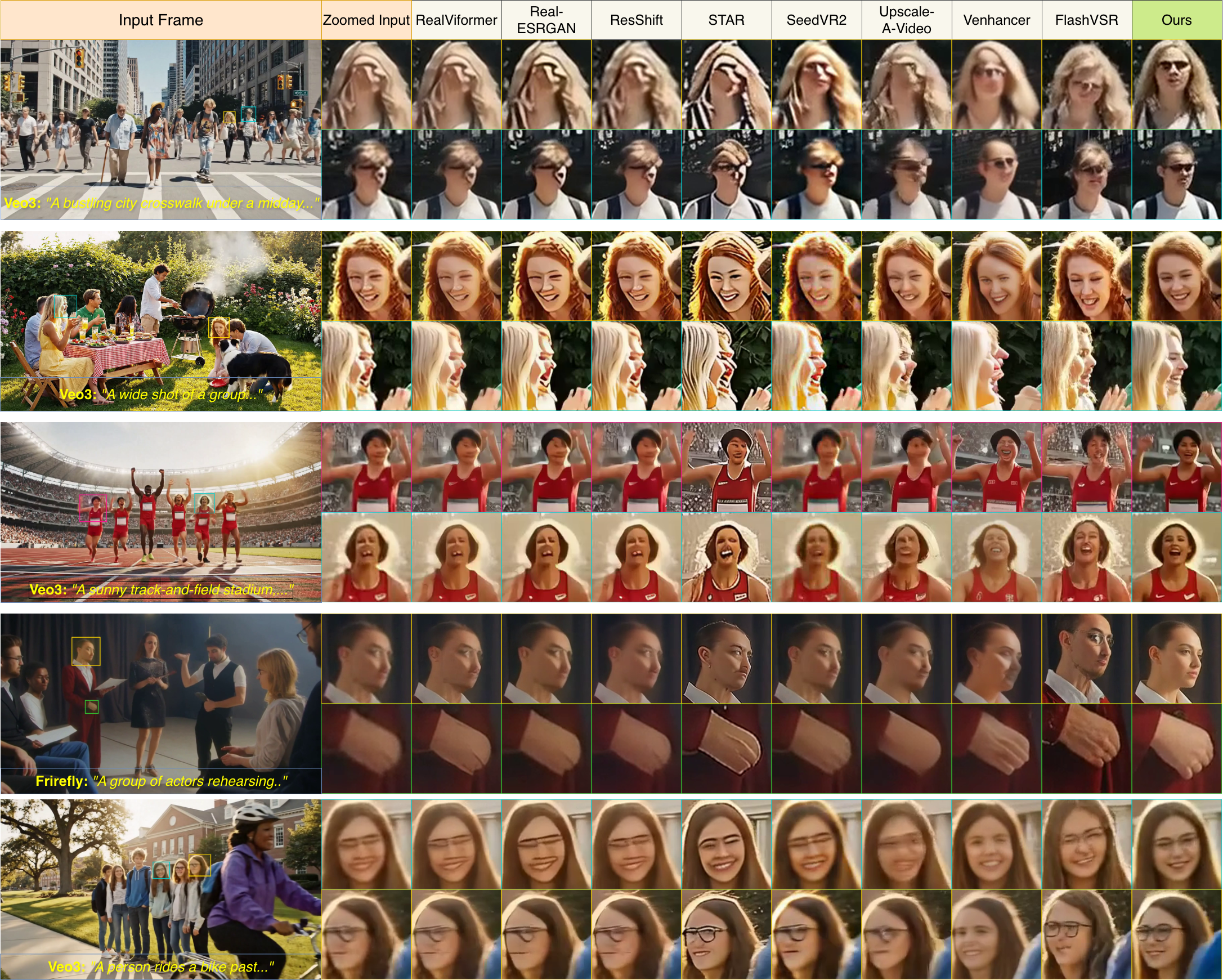}
    \caption{Qualitative Results. We comprehensively compare our method against a wide range of video restoration competitors.} 
    \label{fig:AIGC_Qual1}
\end{figure*}
\subsection{Robustness to input degradations.}
To assess whether the refiner can also operate as a precise video restoration
model, we apply four levels of synthetic input corruption to the REDS30 and
YouHQ40 benchmarks: purely spatial $4\times$ downscaling, joint spatial
$4\times$ + temporal $2\times$ downsampling, and two spatio\mbox{–}temporal
degradation settings Light and Strong defined in ~\Cref{sec:synthetic_degradation}.
We then reconstruct the clean videos and evaluate using precision-oriented
metrics (e.g., PSNR, SSIM, LPIPS, DISTS, NIQE, MUSIQ, CLIP-IQA, and DOVER),
with results reported in \Cref{tab:corruption_reds30,tab:corruption_youhq}.
Across both datasets, our Medium setting already matches or surpasses the best
prior methods on most full-reference metrics for the spatial and mild
spatio\mbox{–}temporal corruptions, while the Strong setting remains competitive
and often leads under the most severe ST-Strong case. This consistent behavior
across four corruption types indicates that the same refiner can robustly undo
both synthetic spatial degradations and harder spatio\mbox{–}temporal artifacts
without sacrificing perceptual quality.

\section{Additional Visualizations}
\label{suppsec:visualizations}
\subsection{Additional qualitative comparisons}
\begin{figure*}[t]
    \centering
    \includegraphics[width=\linewidth]{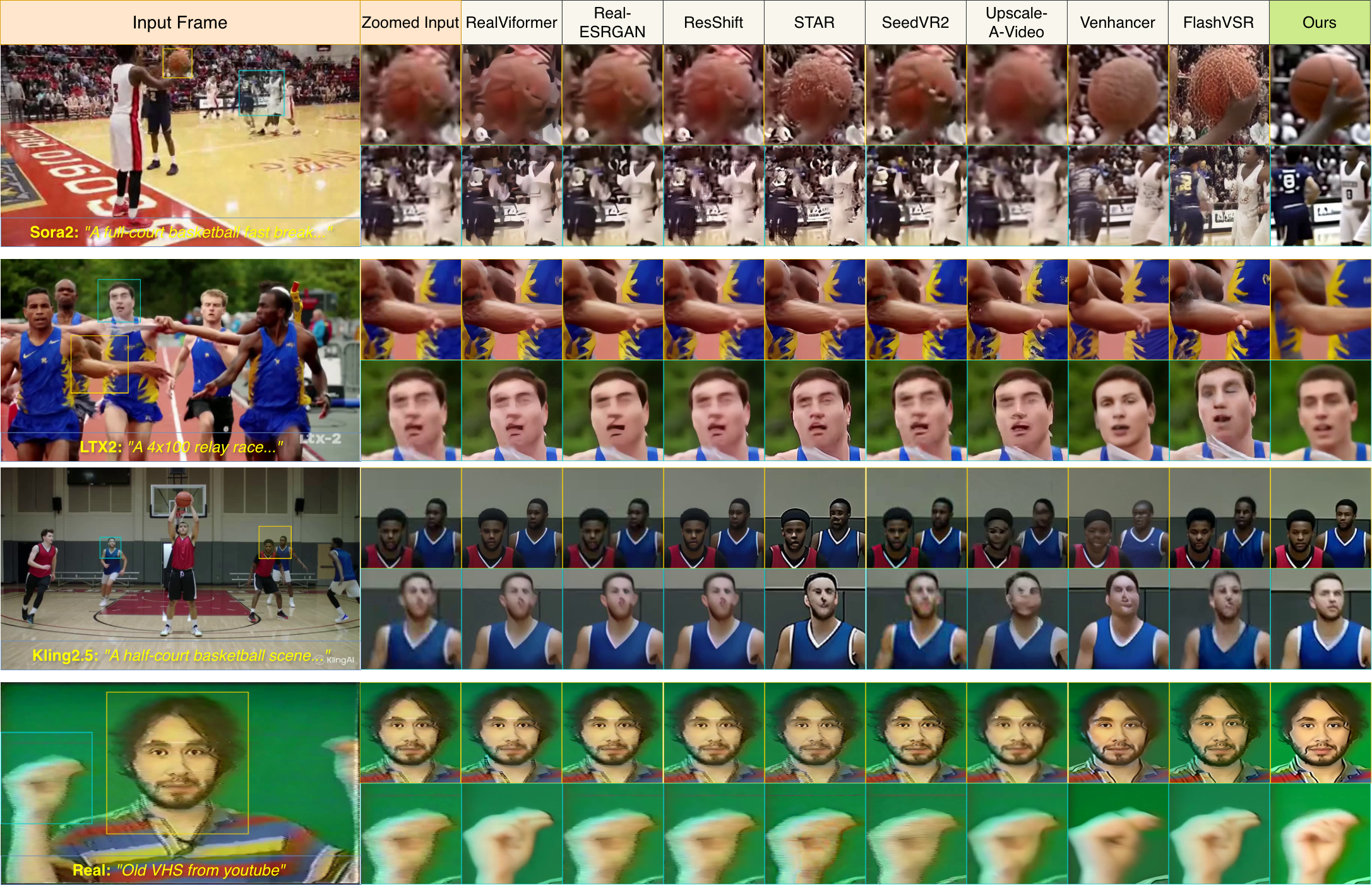}
    \caption{Qualitative Results. We comprehensively compare our method against a wide range of video restoration competitors.} 
    \label{fig:AIGC_Qual2}
\end{figure*}

We provide further qualitative examples from the AIGC benchmark in
\Cref{fig:AIGC_Qual1,fig:AIGC_Qual2}, showing restorations from all competing
methods. Across diverse prompts and source generators, our approach visibly
outperforms prior methods like FlashVSR, Real-Viformer, ResShift, SeedVR2, STAR,
Upscale-A-Video, and VEnhancer: it recovers clearer object and face structures,
suppresses characteristic artifacts, and
produces more natural overall aesthetics, yielding videos that better preserve the intended content.

\subsection{Ablation with Precision Knob during Sampling}
\begin{figure*}[t]
    \centering
    \includegraphics[width=\linewidth]{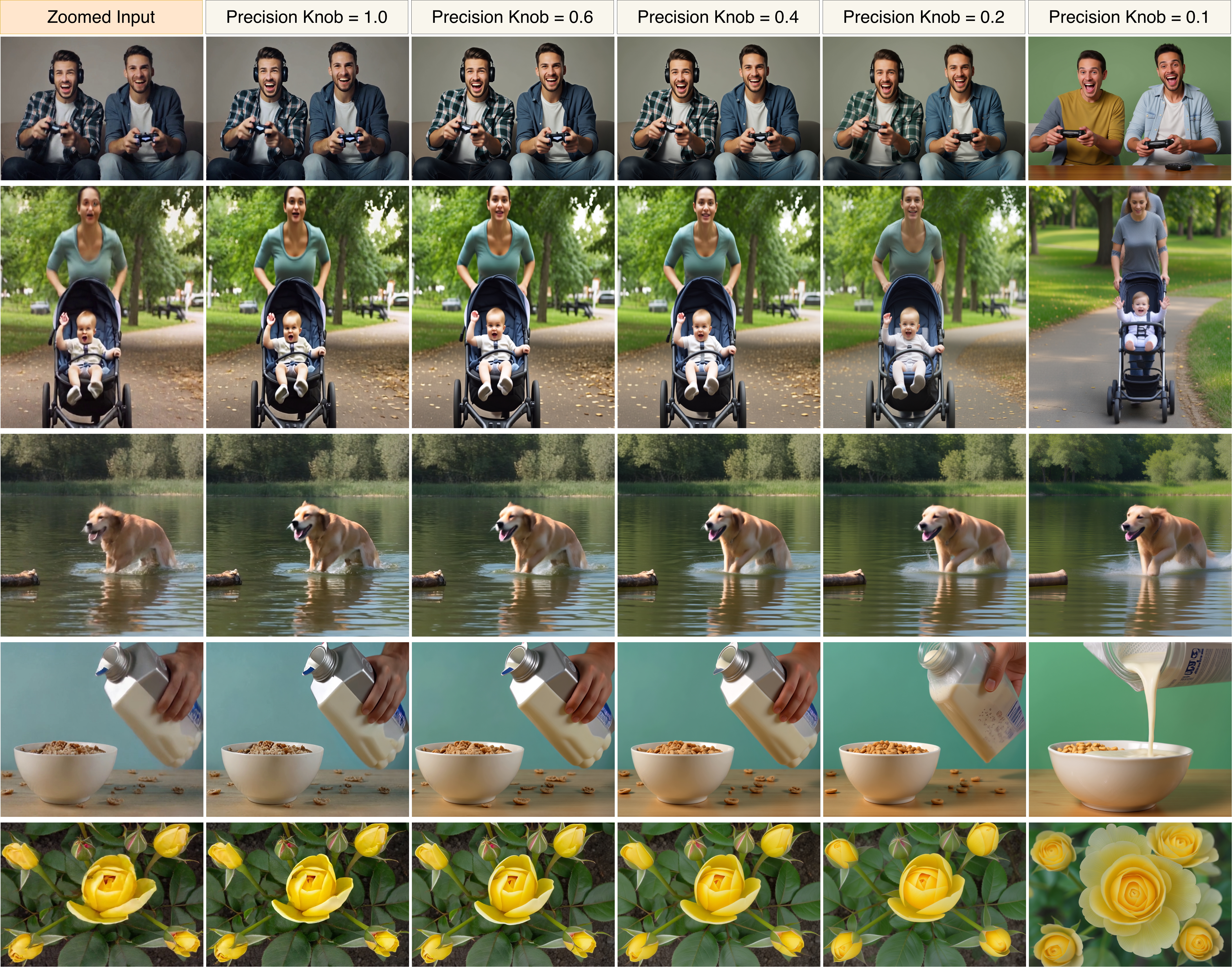}
   \caption{\textbf{Inference Precision knob.} High precision preserves input details; low precision enables stronger corrective synthesis.}
    \label{fig:creative_ablation_supp}
\end{figure*}

At test time we expose a per-layer control scale $\gamma_\ell$ that rescales all
adapter gains, modulating how strongly the degraded video steers the frozen
prior. Larger $\gamma_\ell$ values bias the model toward precise, high-fidelity
restoration that closely follows the input, while smaller $\gamma_\ell$
increase the influence of the prior and enable more aggressive corrective
synthesis. \Cref{fig:creative_ablation_supp} visualizes this trade-off across different
settings of $\gamma_\ell$ for different input samples.

\subsection{Diverse Applications of CreativeVR}
\begin{figure*}[t]
    \centering
    \includegraphics[width=\linewidth]{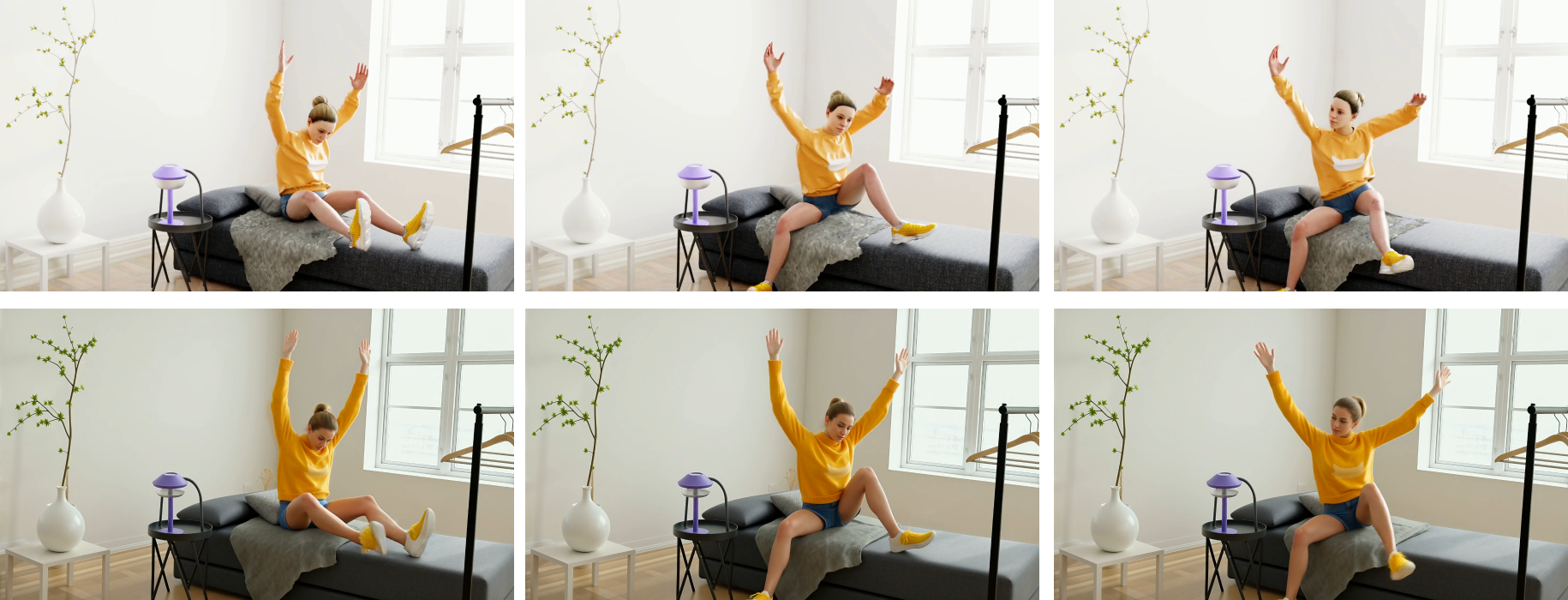}
    \caption{\textbf{CG-to-real translation with CreativeVR.}
    \textbf{Top:} input CG video frames. \textbf{Bottom:} outputs from our
    refiner at a low $\gamma_\ell$, which enhance realism while preserving the
    original character and scene composition.}
    \label{fig:cgtoreal}
\end{figure*}

\paragraph{CG-to-real translation.}
At lower values of the inference scale $\gamma_\ell$, our CreativeVR refiner
can also act as a CG-to-real translator, converting stylized computer-generated
footage into more realistic renderings, as illustrated in
\Cref{fig:cgtoreal}. The identity of the CG character and the overall scene
layout are preserved, while textures, lighting, and shading become more
photorealistic. Higher-resolution visualizations are provided in the attached
video examples.

\begin{figure*}[t]
    \centering
    \includegraphics[width=\linewidth]{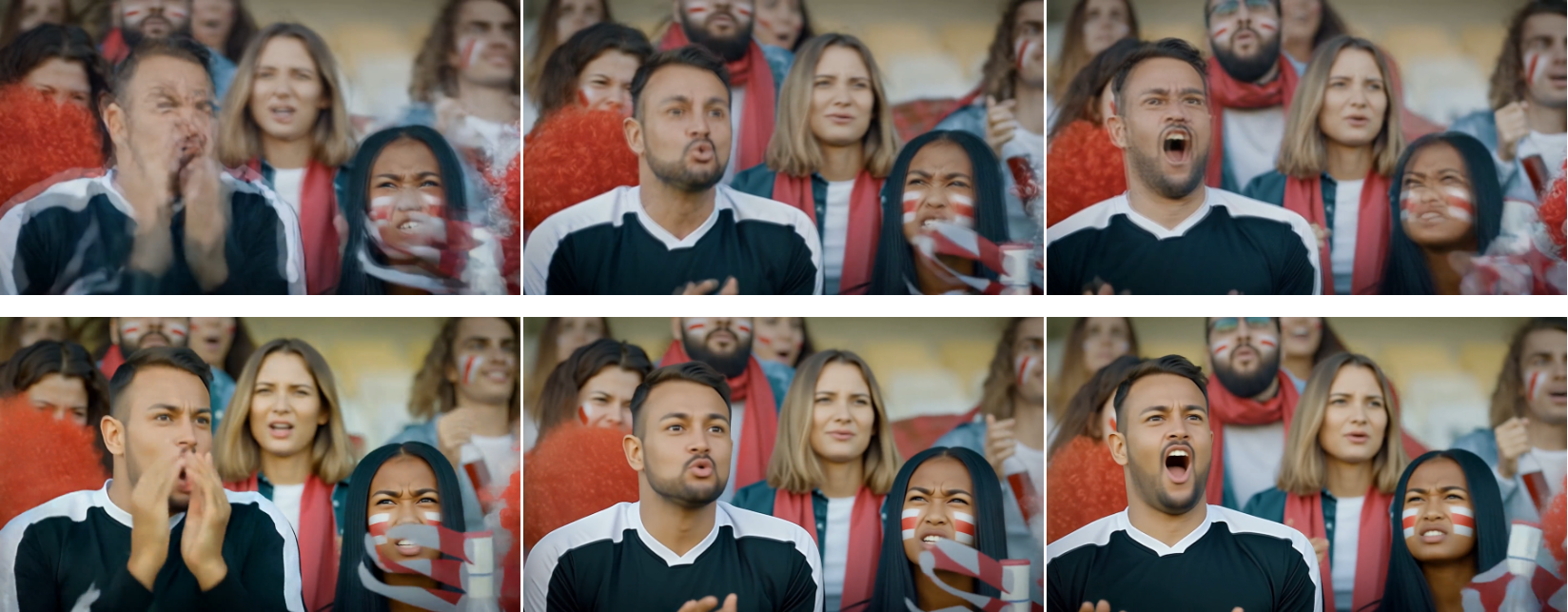}
    \caption{\textbf{Slow-motion generation with CreativeVR.}
\textbf{Top:} linearly interpolated frames used to upsample the input video in
time, which exhibit noticeable morphing and blur artifacts.
\textbf{Bottom:} refined output from CreativeVR, showing sharper details and
more temporally consistent motion, resulting in a high–frame-rate slow-motion
sequence.}
    \label{fig:slowmo}
\end{figure*}

\paragraph{Slow-motion generation.}
Beyond correcting artifacts, CreativeVR can also be used to synthesize
slow-motion, high–frame-rate versions of existing videos. We first upsample
the input sequence in time using simple linear interpolation, which introduces
morphing and blur artifacts in the intermediate frames (top row of
\Cref{fig:slowmo}). Passing this interpolated clip through our refiner yields
temporally smooth, detail-preserving frames (bottom row), effectively turning
the original footage into a visually coherent slow-motion video. This suggests
that CreativeVR can serve as a general frame-rate enhancement module for
arbitrary input videos. Videos are attached in the supplementary. 

Our method also enhances temporal stability by correcting individual frames using the overall temporal context; corresponding examples are provided in the attached videos under \emph{Applications}.

\clearpage
\balance
{\small
\bibliographystyle{ieee_fullname}
\bibliography{egbib}

@String(PAMI = {IEEE Trans. Pattern Anal. Mach. Intell.})

@String(IJCV = {Int. J. Comput. Vis.})

@String(CVPR= {IEEE Conf. Comput. Vis. Pattern Recog.})

@String(ICPR = {Int. Conf. Pattern Recog.})

@String(ICLR = {Int. Conf. Learn. Represent.})

@String(PAMI  = {IEEE TPAMI})

@String(IJCV  = {IJCV})

@String(CVPR  = {CVPR})

@String(ICPR  = {ICPR})

@String(ICLR  = {ICLR})

@misc{he2024venhancer,
  author = {He, J. and Xue, T. and Liu, D. and Lin, X. and Gao, P. and Lin, D. and Qiao, Y. and Ouyang, W. and Liu, Z.},
  title = {VEnhancer: Generative Space-Time Enhancement for Video Generation},
  howpublished = {OpenReview (ICLR 2025 submission)},
  year = {2024},
  note = {OpenReview / ICLR submission (see provided OpenReview link)}
}

@misc{xu2024videogigagan,
  author = {Xu, Y. and Park, T. and Zhang, R. and Zhou, Y. and Shechtman, E. and Liu, F. and Huang, J.-B. and Liu, D.},
  title = {VideoGigaGAN: Towards Detail-rich Video Super-Resolution},
  year = {2024},
  howpublished = {arXiv preprint},
  eprint = {arXiv:2404.12388}
}

@inproceedings{wang2019edvr,
  author = {Wang, Xintao and Chan, Kelvin C. K. and Yu, Ke and Dong, Chao and Loy, Chen Change},
  title = {EDVR: Video Restoration with Enhanced Deformable Convolutions},
  booktitle = {Proceedings of the CVPR Workshops (NTIRE) 2019},
  year = {2019},
  url = {https://openaccess.thecvf.com/content_CVPRW_2019/papers/NTIRE/Wang_EDVR_Video_Restoration_With_Enhanced_Deformable_Convolutional_Networks_CVPRW_2019_paper.pdf}
}

@inproceedings{chan2022basicvsrplusplus,
  author = {Chan, Kelvin C.K. and Zhou, Shiyu and Xu, Xintao and Loy, Chen Change},
  title = {BasicVSR++: Improving Video Super-Resolution with Enhanced Propagation and Alignment},
  booktitle = {Proc. CVPR},
  year = {2022},
  pages = {XXX--XXX},
  url = {https://openaccess.thecvf.com/content/CVPR2022/papers/Chan_BasicVSR_Improving_Video_Super-Resolution_With_Enhanced_Propagation_and_Alignment_CVPR_2022_paper.pdf}
}

@misc{wang2025seedvr,
  author = {Wang, J. and Lin, Z. and Wei, M. and Zhao, Y. and Loy, C. C. and Jiang, L. and Yang, C.},
  title = {SeedVR: Seeding Infinity in Diffusion Transformer Towards Generic Video Restoration},
  year = {2025},
  howpublished = {arXiv preprint},
  eprint = {arXiv:2501.01320}
}

@misc{wang2024stablesr,
  author = {Wang, J. and Yue, Z. and Zhou, S. and Chan, K. C. K. and Loy, C. C.},
  title = {Exploiting Diffusion Prior for Real-World Image Super-Resolution (StableSR)},
  year = {2024},
  howpublished = {IJCV (preprint) / arXiv},
  eprint = {arXiv:2305.07015}
}

@article{houlsby2019adapters,
  author = {Houlsby, Neil and Giurgiu, Andrei and Jastrzebski, Stanislaw and Morrone, Ben and De Laroussilhe, Quentin and Gesmundo, Antonio and Attariyan, Moe and Gelly, Sylvain},
  title = {Parameter-Efficient Transfer Learning for NLP},
  journal = {ICML Workshop / arXiv},
  year = {2019},
  eprint = {arXiv:1902.00751}
}

@misc{hu2021lora,
  author = {Hu, Edward J. and Shen, Yelong and Wallis, Phillip and Allen-Zhu, Zeyuan and Li, Yuanzhi and Wang, Shean and Chen, Weizhu},
  title = {LoRA: Low-Rank Adaptation of Large Language Models},
  year = {2021},
  howpublished = {arXiv preprint},
  eprint = {arXiv:2106.09685}
}

@misc{xu2024upscaleavideo,
  author = {Zhou, P. and others},
  title = {Upscale-A-Video: Text-Guided Latent Diffusion for Video Upscaling},
  year = {2024},
  howpublished = {arXiv preprint},
  note = {Text-guided upscaling / prompt-driven texture synthesis; replace with preferred paper details}
}

@misc{xie2025star,
  author = {Xie, R. and Liu, Y. and Zhou, P. and Zhao, C. and Zhou, J. and Zhang, K. and Zhang, Z. and Yang, J. and Yang, Z. and Tai, Y.},
  title = {STAR: Spatial-Temporal Augmentation with Text-to-Video Models for Real-World Video Super-Resolution},
  year = {2025},
  howpublished = {arXiv preprint},
  eprint = {arXiv:2501.02976}
}

@inproceedings{chan2022basicvsr,
  title={Basicvsr++: Improving video super-resolution with enhanced propagation and alignment},
  author={Chan, Kelvin CK and Zhou, Shangchen and Xu, Xiangyu and Loy, Chen Change},
  booktitle={Proceedings of the IEEE/CVF conference on computer vision and pattern recognition},
  pages={5972--5981},
  year={2022}
}

@inproceedings{lugmayr2022repaint,
  title={Repaint: Inpainting using denoising diffusion probabilistic models},
  author={Lugmayr, Andreas and Danelljan, Martin and Romero, Andres and Yu, Fisher and Timofte, Radu and Van Gool, Luc},
  booktitle={Proceedings of the IEEE/CVF conference on computer vision and pattern recognition},
  pages={11461--11471},
  year={2022}
}

@article{wang2025seedvr2,
  title={Seedvr2: One-step video restoration via diffusion adversarial post-training},
  author={Wang, Jianyi and Lin, Shanchuan and Lin, Zhijie and Ren, Yuxi and Wei, Meng and Yue, Zongsheng and Zhou, Shangchen and Chen, Hao and Zhao, Yang and Yang, Ceyuan and others},
  journal={arXiv preprint arXiv:2506.05301},
  year={2025}
}

@article{zhang2025infvsr,
  title={InfVSR: Breaking Length Limits of Generic Video Super-Resolution},
  author={Zhang, Ziqing and Liu, Kai and Chen, Zheng and Li, Xi and Chen, Yucong and Duan, Bingnan and Kong, Linghe and Zhang, Yulun},
  journal={arXiv preprint arXiv:2510.00948},
  year={2025}
}

@article{zhang2025flashvideo,
  title={Flashvideo: Flowing fidelity to detail for efficient high-resolution video generation},
  author={Zhang, Shilong and Li, Wenbo and Chen, Shoufa and Ge, Chongjian and Sun, Peize and Zhang, Yida and Jiang, Yi and Yuan, Zehuan and Peng, Binyue and Luo, Ping},
  journal={arXiv preprint arXiv:2502.05179},
  year={2025}
}

@article{jiang2025vace,
  title={Vace: All-in-one video creation and editing},
  author={Jiang, Zeyinzi and Han, Zhen and Mao, Chaojie and Zhang, Jingfeng and Pan, Yulin and Liu, Yu},
  journal={arXiv preprint arXiv:2503.07598},
  year={2025}
}

@article{jiang2023res,
  title={Res-tuning: A flexible and efficient tuning paradigm via unbinding tuner from backbone},
  author={Jiang, Zeyinzi and Mao, Chaojie and Huang, Ziyuan and Ma, Ao and Lv, Yiliang and Shen, Yujun and Zhao, Deli and Zhou, Jingren},
  journal={Advances in Neural Information Processing Systems},
  volume={36},
  pages={42689--42716},
  year={2023}
}

@article{zhuang2025flashvsr,
  title={FlashVSR: Towards Real-Time Diffusion-Based Streaming Video Super-Resolution},
  author={Zhuang, Junhao and Guo, Shi and Cai, Xin and Li, Xiaohui and Liu, Yihao and Yuan, Chun and Xue, Tianfan},
  journal={arXiv preprint arXiv:2510.12747},
  year={2025}
}

@inproceedings{zhang2024realviformer,
  title={Realviformer: Investigating attention for real-world video super-resolution},
  author={Zhang, Yuehan and Yao, Angela},
  booktitle={European Conference on Computer Vision},
  pages={412--428},
  year={2024},
  organization={Springer}
}

@article{yue2023resshift,
  title={Resshift: Efficient diffusion model for image super-resolution by residual shifting},
  author={Yue, Zongsheng and Wang, Jianyi and Loy, Chen Change},
  journal={Advances in Neural Information Processing Systems},
  volume={36},
  pages={13294--13307},
  year={2023}
}

@inproceedings{wang2021real,
  title={Real-esrgan: Training real-world blind super-resolution with pure synthetic data},
  author={Wang, Xintao and Xie, Liangbin and Dong, Chao and Shan, Ying},
  booktitle={Proceedings of the IEEE/CVF international conference on computer vision},
  pages={1905--1914},
  year={2021}
}

@inproceedings{yi2019progressive,
  title={Progressive fusion video super-resolution network via exploiting non-local spatio-temporal correlations},
  author={Yi, Peng and Wang, Zhongyuan and Jiang, Kui and Jiang, Junjun and Ma, Jiayi},
  booktitle={Proceedings of the IEEE/CVF international conference on computer vision},
  pages={3106--3115},
  year={2019}
}

@inproceedings{tao2017detail,
  title={Detail-revealing deep video super-resolution},
  author={Tao, Xin and Gao, Hongyun and Liao, Renjie and Wang, Jue and Jia, Jiaya},
  booktitle={Proceedings of the IEEE international conference on computer vision},
  pages={4472--4480},
  year={2017}
}

@inproceedings{nah2019ntire,
  title={Ntire 2019 challenge on video deblurring and super-resolution: Dataset and study},
  author={Nah, Seungjun and Baik, Sungyong and Hong, Seokil and Moon, Gyeongsik and Son, Sanghyun and Timofte, Radu and Mu Lee, Kyoung},
  booktitle={Proceedings of the IEEE/CVF conference on computer vision and pattern recognition workshops},
  pages={0--0},
  year={2019}
}

@article{saharia2022image,
  title={Image super-resolution via iterative refinement},
  author={Saharia, Chitwan and Ho, Jonathan and Chan, William and Salimans, Tim and Fleet, David J and Norouzi, Mohammad},
  journal={IEEE transactions on pattern analysis and machine intelligence},
  volume={45},
  number={4},
  pages={4713--4726},
  year={2022},
  publisher={IEEE}
}

@article{luo2023image,
  title={Image restoration with mean-reverting stochastic differential equations},
  author={Luo, Ziwei and Gustafsson, Fredrik K and Zhao, Zheng and Sj{\"o}lund, Jens and Sch{\"o}n, Thomas B},
  journal={arXiv preprint arXiv:2301.11699},
  year={2023}
}

@inproceedings{zhang2021designing,
  title={Designing a practical degradation model for deep blind image super-resolution},
  author={Zhang, Kai and Liang, Jingyun and Van Gool, Luc and Timofte, Radu},
  booktitle={Proceedings of the IEEE/CVF international conference on computer vision},
  pages={4791--4800},
  year={2021}
}

@article{wang2024exploiting,
  title={Exploiting diffusion prior for real-world image super-resolution},
  author={Wang, Jianyi and Yue, Zongsheng and Zhou, Shangchen and Chan, Kelvin CK and Loy, Chen Change},
  journal={International Journal of Computer Vision},
  volume={132},
  number={12},
  pages={5929--5949},
  year={2024},
  publisher={Springer}
}

@article{li2025diffusion,
  title={Diffusion models for image restoration and enhancement: a comprehensive survey},
  author={Li, Xin and Ren, Yulin and Jin, Xin and Lan, Cuiling and Wang, Xingrui and Zeng, Wenjun and Wang, Xinchao and Chen, Zhibo},
  journal={International Journal of Computer Vision},
  pages={1--31},
  year={2025},
  publisher={Springer}
}

@inproceedings{kong2025deblurdiff,
  title={DeblurDiff: Real-Word Image Deblurring with Generative Diffusion Models},
  author={Kong, Lingshun and Zou, Dongqing and Wang, Fu Lee and Ren, Jimmy and Wu, Xiaohe and Dong, Jiangxin and Pan, Jinshan and others},
  booktitle={The Thirty-ninth Annual Conference on Neural Information Processing Systems},
  year={2025}
}

@inproceedings{tian2020tdan,
  title={Tdan: Temporally-deformable alignment network for video super-resolution},
  author={Tian, Yapeng and Zhang, Yulun and Fu, Yun and Xu, Chenliang},
  booktitle={Proceedings of the IEEE/CVF conference on computer vision and pattern recognition},
  pages={3360--3369},
  year={2020}
}

@inproceedings{fuoli2019efficient,
  title={Efficient video super-resolution through recurrent latent space propagation},
  author={Fuoli, Dario and Gu, Shuhang and Timofte, Radu},
  booktitle={2019 IEEE/CVF International Conference on Computer Vision Workshop (ICCVW)},
  pages={3476--3485},
  year={2019},
  organization={IEEE}
}

@inproceedings{xu2021temporal,
  title={Temporal modulation network for controllable space-time video super-resolution},
  author={Xu, Gang and Xu, Jun and Li, Zhen and Wang, Liang and Sun, Xing and Cheng, Ming-Ming},
  booktitle={Proceedings of the IEEE/CVF conference on computer vision and pattern recognition},
  pages={6388--6397},
  year={2021}
}

@inproceedings{kang2023scaling,
  title={Scaling up gans for text-to-image synthesis},
  author={Kang, Minguk and Zhu, Jun-Yan and Zhang, Richard and Park, Jaesik and Shechtman, Eli and Paris, Sylvain and Park, Taesung},
  booktitle={Proceedings of the IEEE/CVF conference on computer vision and pattern recognition},
  pages={10124--10134},
  year={2023}
}

@inproceedings{xu2025videogigagan,
  title={Videogigagan: Towards detail-rich video super-resolution},
  author={Xu, Yiran and Park, Taesung and Zhang, Richard and Zhou, Yang and Shechtman, Eli and Liu, Feng and Huang, Jia-Bin and Liu, Difan},
  booktitle={Proceedings of the Computer Vision and Pattern Recognition Conference},
  pages={2139--2149},
  year={2025}
}

@inproceedings{zhang2023adding,
  title={Adding conditional control to text-to-image diffusion models},
  author={Zhang, Lvmin and Rao, Anyi and Agrawala, Maneesh},
  booktitle={Proceedings of the IEEE/CVF international conference on computer vision},
  pages={3836--3847},
  year={2023}
}

@article{ju2025editverse,
  title={EditVerse: Unifying Image and Video Editing and Generation with In-Context Learning},
  author={Ju, Xuan and Wang, Tianyu and Zhou, Yuqian and Zhang, He and Liu, Qing and Zhao, Nanxuan and Zhang, Zhifei and Li, Yijun and Cai, Yuanhao and Liu, Shaoteng and others},
  journal={arXiv preprint arXiv:2509.20360},
  year={2025}
}

@inproceedings{yin2025slow,
  title={From slow bidirectional to fast autoregressive video diffusion models},
  author={Yin, Tianwei and Zhang, Qiang and Zhang, Richard and Freeman, William T and Durand, Fredo and Shechtman, Eli and Huang, Xun},
  booktitle={Proceedings of the Computer Vision and Pattern Recognition Conference},
  pages={22963--22974},
  year={2025}
}

@article{salimans2024multistep,
  title={Multistep distillation of diffusion models via moment matching},
  author={Salimans, Tim and Mensink, Thomas and Heek, Jonathan and Hoogeboom, Emiel},
  journal={Advances in Neural Information Processing Systems},
  volume={37},
  pages={36046--36070},
  year={2024}
}

@article{huang2025self,
  title={Self Forcing: Bridging the Train-Test Gap in Autoregressive Video Diffusion},
  author={Huang, Xun and Li, Zhengqi and He, Guande and Zhou, Mingyuan and Shechtman, Eli},
  journal={arXiv preprint arXiv:2506.08009},
  year={2025}
}

@article{luo2023latent,
  title={Latent consistency models: Synthesizing high-resolution images with few-step inference},
  author={Luo, Simian and Tan, Yiqin and Huang, Longbo and Li, Jian and Zhao, Hang},
  journal={arXiv preprint arXiv:2310.04378},
  year={2023}
}

@article{babnik2024ediffiqa,
  title={eDifFIQA: towards efficient face image quality assessment based on denoising diffusion probabilistic models},
  author={Babnik, {\v{Z}}iga and Peer, Peter and {\v{S}}truc, Vitomir},
  journal={IEEE Transactions on Biometrics, Behavior, and Identity Science},
  volume={6},
  number={4},
  pages={458--474},
  year={2024},
  publisher={IEEE}
}

@inproceedings{babnik2023diffiqa,
  title={DifFIQA: Face image quality assessment using denoising diffusion probabilistic models},
  author={Babnik, {\v{Z}}iga and Peer, Peter and {\v{S}}truc, Vitomir},
  booktitle={2023 IEEE international joint conference on biometrics (IJCB)},
  pages={1--10},
  year={2023},
  organization={IEEE}
}

@inproceedings{ou2024clib,
  title={Clib-fiqa: Face image quality assessment with confidence calibration},
  author={Ou, Fu-Zhao and Li, Chongyi and Wang, Shiqi and Kwong, Sam},
  booktitle={Proceedings of the IEEE/CVF Conference on Computer Vision and Pattern Recognition},
  pages={1694--1704},
  year={2024}
}

@inproceedings{boutros2023cr,
  title={CR-FIQA: face image quality assessment by learning sample relative classifiability},
  author={Boutros, Fadi and Fang, Meiling and Klemt, Marcel and Fu, Biying and Damer, Naser},
  booktitle={Proceedings of the IEEE/CVF conference on computer vision and pattern recognition},
  pages={5836--5845},
  year={2023}
}

@inproceedings{ou2025mr,
  title={MR-FIQA: Face Image Quality Assessment with Multi-Reference Representations from Synthetic Data Generation},
  author={Ou, Fu-Zhao and Li, Chongyi and Wang, Shiqi and Kwong, Sam},
  booktitle={Proceedings of the IEEE/CVF International Conference on Computer Vision},
  pages={12915--12925},
  year={2025}
}

@inproceedings{babnik2022faceqan,
  title={FaceQAN: Face image quality assessment through adversarial noise exploration},
  author={Babnik, {\v{Z}}iga and Peer, Peter and {\v{S}}truc, Vitomir},
  booktitle={2022 26th International Conference on Pattern Recognition (ICPR)},
  pages={748--754},
  year={2022},
  organization={IEEE}
}

@article{opensoraplan2024,
  title   = {Open-Sora Plan: Open-Source Large Video Generation Model},
  author  = {{Open-Sora Plan Team}},
  journal = {arXiv preprint arXiv:2412.00131},
  year    = {2024}
}

@inproceedings{lpips,
  title     = {The Unreasonable Effectiveness of Deep Features as a Perceptual Metric},
  author    = {Zhang, Richard and Isola, Phillip and Efros, Alexei A. and Shechtman, Eli and Wang, Oliver},
  booktitle = CVPR,
  pages     = {586--595},
  year      = {2018}
}

@article{dists,
  title   = {Image Quality Assessment: Unifying Structure and Texture Similarity},
  author  = {Ding, Keyan and Ma, Kede and Wang, Shiqi and Simoncelli, Eero P.},
  journal = PAMI,
  volume  = {44},
  number  = {5},
  pages   = {2567--2581},
  year    = {2022}
}

@article{ssim,
  title   = {Image Quality Assessment: From Error Visibility to Structural Similarity},
  author  = {Wang, Zhou and Bovik, Alan C. and Sheikh, Hamid R. and Simoncelli, Eero P.},
  journal = PAMI,
  volume  = {26},
  number  = {4},
  pages   = {600--612},
  year    = {2004}
}

@inproceedings{psnr,
  title     = {Image Quality Metrics: PSNR vs. SSIM},
  author    = {Hore, Artur and Ziou, Dimitrios},
  booktitle = ICPR,
  pages     = {2366--2369},
  year      = {2010}
}

@misc{google2024veo3,
  title        = {Veo 3},
  author       = {Google DeepMind},
  year         = {2024},
  howpublished = {\url{https://deepmind.google/technologies/veo/}},
  note         = {Accessed: 2025-01}
}

@misc{luma2025ray3,
  title        = {Ray3: Intelligent Video Model with HDR and Visual Reasoning},
  author       = {Luma AI},
  year         = {2025},
  howpublished = {\url{https://lumalabs.ai/ray}},
  note         = {Accessed: 2025-11-13}
}

@misc{openai2025sora2,
  title        = {Sora 2},
  author       = {OpenAI},
  year         = {2025},
  howpublished = {\url{https://openai.com/sora}},
  note         = {Accessed: 2025-11-13}
}

@article{gao2025seedance,
  title={Seedance 1.0: Exploring the Boundaries of Video Generation Models},
  author={Gao, Yu and Guo, Haoyuan and Hoang, Tuyen and Huang, Weilin and Jiang, Lu and Kong, Fangyuan and Li, Huixia and Li, Jiashi and Li, Liang and Li, Xiaojie and others},
  journal={arXiv preprint arXiv:2506.09113},
  year={2025}
}

@misc{pika2025pika22,
  title        = {Pika 2.2},
  author       = {Pika Labs},
  year         = {2025},
  howpublished = {\url{https://pika.art}},
  note         = {Accessed: 2025-11-13}
}

@misc{adobe2025firefly,
  title        = {Adobe Firefly},
  author       = {Adobe},
  year         = {2025},
  howpublished = {\url{https://www.adobe.com/sensei/generative-ai/firefly.html}},
  note         = {Accessed: 2025-11-13}
}

@misc{ltx2_2025,
  title        = {LTX-2: The Next-Generation Multimodal AI Video Foundation Model},
  author       = {{Lightricks}},
  year         = {2025},
  howpublished = {\url{https://ltx.video/}},
  note         = {Accessed: 2025-11-14}
}

@article{wan2025wan,
  title={Wan: Open and advanced large-scale video generative models},
  author={Wan, Team and Wang, Ang and Ai, Baole and Wen, Bin and Mao, Chaojie and Xie, Chen-Wei and Chen, Di and Yu, Feiwu and Zhao, Haiming and Yang, Jianxiao and others},
  journal={arXiv preprint arXiv:2503.20314},
  year={2025}
}

@misc{kling2025texttovideo,
  title        = {Kling: A Text-to-Video Generation Model},
  author       = {Kuaishou Technology},
  year         = {2025},
  howpublished = {\url{https://klingai.com/}},  
  note         = {Accessed: 2025-11-14}
}

@inproceedings{huang2024vbench,
  title={Vbench: Comprehensive benchmark suite for video generative models},
  author={Huang, Ziqi and He, Yinan and Yu, Jiashuo and Zhang, Fan and Si, Chenyang and Jiang, Yuming and Zhang, Yuanhan and Wu, Tianxing and Jin, Qingyang and Chanpaisit, Nattapol and others},
  booktitle={Proceedings of the IEEE/CVF Conference on Computer Vision and Pattern Recognition},
  pages={21807--21818},
  year={2024}
}
}

\end{document}